\documentclass[onecolumn]{IEEEtran}

\usepackage{cite}
\usepackage{amsmath, amssymb, amsfonts}
\usepackage{algorithmic}
\usepackage[ruled,vlined]{algorithm2e}
\usepackage{xcolor}
\usepackage{hyperref}
\usepackage{graphicx}
\usepackage{textcomp}
\usepackage{subfig}
\usepackage{multirow}
\usepackage{svg}
\usepackage{colortbl}
\usepackage{soul}
\usepackage{enumerate}

\usepackage{comment}

\usepackage{hyperref}

\def\BibTeX{{\rm B\kern-.05em{\sc i\kern-.025em b}\kern-.08em
    T\kern-.1667em\lower.7ex\hbox{E}\kern-.125emX}}
\begin{document}

\author{\IEEEauthorblockN{Maloy Kumar Devnath, Avijoy Chakma, Mohammad Saeid Anwar, Emon Dey, Zahid Hasan, Marc Conn*, \\ Biplab Pal, Nirmalya Roy}\\
\vspace{0.3 em}
\IEEEauthorblockA{\textit{Information Systems Department, *Computer Engineering Department\\}}
\IEEEauthorblockA{University of Maryland, Baltimore County (UMBC), USA\\}
\IEEEauthorblockA{\{maloyd1, achakma1, saeid.anwar, edey1, zhasan3, mconn1, bpal1, nroy\}@umbc.edu}
}




\title{A Systematic Study on Object Recognition Using Millimeter-wave Radar
\thanks{ 
\textit{Corresponding author: Maloy Kumar Devnath}}
}

\maketitle

\begin{abstract}

Millimeter-wave (MMW) radar is becoming an essential sensing technology in smart environments due to its light and weather-independent sensing capability. Such capabilities have been widely explored and integrated with intelligent vehicle systems, often deployed in industry-grade MMW radars. However, industry-grade MMW radars are often expensive and difficult to attain for deployable community-purpose smart environment applications. On the other hand, commercially available MMW radars pose hidden underpinning challenges that are yet to be well investigated for tasks such as recognizing objects, and activities, real-time person tracking, object localization, etc. Such tasks are frequently accompanied by image and video data, which are relatively easy for an individual to obtain, interpret, and annotate. However, image and video data are light and weather-dependent, vulnerable to the occlusion effect, and inherently raise privacy concerns for individuals. It is crucial to investigate the performance of an alternative sensing mechanism where commercially available MMW radars can be a viable alternative to eradicate the dependencies and preserve privacy issues. Before championing MMW radar, several questions need to be answered regarding MMW radar's practical feasibility and performance under different operating environments. To answer the concerns, we have collected a dataset using commercially available MMW radar, Automotive mmWave Radar (AWR2944) from Texas Instruments, and reported the optimum experimental settings for object recognition performance using several deep learning algorithms in this study. Moreover, our robust data collection procedure allows us to systematically study and identify potential challenges in the object recognition task under a cross-ambience scenario. We have explored the potential approaches to overcome the underlying challenges and reported extensive experimental results.

\end{abstract}



\section{Introduction}

\label{sec:intro}

Visual data, such as images and videos, has been extensively used in building smart environments for various applications, including remote monitoring, security management, object recognition, tracking mobile objects, autonomous agents, and more~\cite{brumitt2000easyliving,dey2022synchrosim}. However, imaging as a sensing mechanism is susceptible to environmental factors such as lighting and weather, and the privacy of individuals may be compromised when captured by cameras. Under certain circumstances, privacy is a non-negotiable constraint, and therefore, the response to the concerns mentioned below related to visual data may not be positive.

\begin{enumerate}
    \item would the tenants in an elderly care home be comfortable and agree to be watched 24 hours by multiple cameras?
    \item how would a camera-based system respond, if a person is lying on the kitchen floor due to sickness at night?
    \item can the tiny hazardous material in a playing grass field be identified by images?
\end{enumerate}

\begin{figure}[!ht]
    \centering
     \includegraphics[width =0.60 \linewidth]{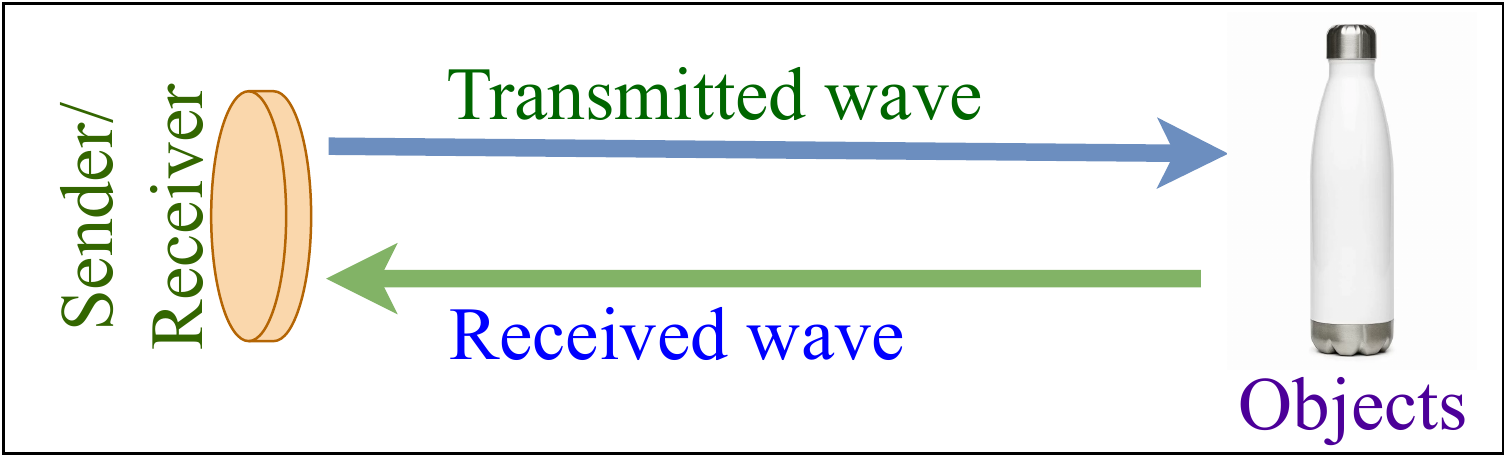}
    \caption{MMW radar's basic principles.}
    \label{fig:mmWave_functionality}
\end{figure}

The above concerns have motivated us to explore an active sensing mechanism such as RADAR-based technology that has the potential to bypass the constraints of privacy and environmental dependencies. Note that, RADAR technology with various capacities has been explored in numerous applications such as remote monitoring~\cite{alizadeh2019}, security management~\cite{dong2021}, object recognition~\cite{wang2021rodnet}, tracking mobile objects~\cite{xu2022object,zhao2019mid}, autonomous agents (unmanned ground vehicle)~\cite{amsters2020turtlebot} and many more.  We find that these radar technologies are often expensive, difficult to develop in a timely manner, and require substantial domain knowledge~\cite{alizadeh2019} which is a deterrent factor for a community-based deployment. MMW radar is a recent addition to the radar technology pool that projects short wavelength (30-300GHz) electromagnetic waves as short pulses and receives the reflected pulses from the objects that lie in its projection path~\cite{iovescu2020fundamentals} as depicted in Figure~\ref{fig:mmWave_functionality}. MMW radar is commercially available and can be integrated with research-grade programmable autonomous agents and vehicles~\cite{green2019using,toker2020mmwave,amsters2020turtlebot,islam2022ggnb}. However, MMW radar is not entirely free of constraints, which leads us to conduct this systematic study and investigate those underpinning challenges.

\begin{figure}[!ht]
    \centering
     \includegraphics[width =0.50 \linewidth]{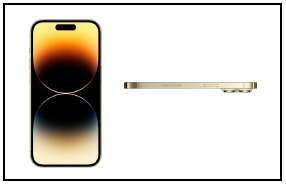}
    \caption{Smart phone's top and side view.}
    \label{fig:smart_phone}
\end{figure}

Most of the literature that has been discussed in Section~\ref{sec:literature} involves large objects~\cite{kosuge2022mmwave} and ignores smaller objects under varying environmental conditions where the next one pose several challenges. The challenges become more severe when we use the sparse point cloud data of the MMW radar. First,~\textbf{Impact of object size and shape} - hazardous objects such as broken metal prongs, pets in a home environment, electronic devices, and lanyards are often small in shape, therefore small surface area to receive and reflect the radar signal. Second,~\textbf{Operating Environment} - often objects are composed of materials with varying reflective characteristics. When radar projects a millimeter wave on an object, the resulting reflected signal waves carry varied signal-to-noise ratios. The reflected signal from the surrounding objects complicates the final received signal and therefore the object recognition task. Third,~\textbf{Radar Position and Distance} - radar position and the relevant distance from the object are another impacting factor. Consider the top and side view of a smartphone as depicted in Figure~\ref{fig:smart_phone} and the change in the surface area is obvious. In addition, as the distance between the radar and the object increases, the projected signal diverges into a wider area, resulting in an attenuation of the received signal. Currently, the object recognition performance of a commercially available MMW radar under varying shapes, environments, and dynamic settings is unknown.

In this study, we make the following contributions -

\begin{enumerate}

    \item We develop models to recognize small-sized objects using low-resolution and sparse point cloud MMW radar data in three indoor and outdoor environments with varying settings, including static or dynamic radar projection and height or distance.

    \item We are using domain adaptation techniques to improve the robustness of a deep learning model on a large scale, specifically for recognizing small objects in low-resolution and sparse point cloud data obtained from MMW radar. To the best of our knowledge, this is the first study to explore domain adaptation for these types of objects.

    \item  We have validated our proposed system by conducting experiments in sunny outdoor, well-lit indoor, and dark indoor settings with a large-scale dataset collected using a TI AWR2944 MMW radar. Particularly, we focus on a feasibility study to recognize varying-sized small objects such as dimes, pencils, plastic sheets, quarters, Unnamed Ground Vehicle (UGV), water bottles, wood, pencils, and paper under varying environmental settings (distance, height, lighting, backgrounds) and collect around five hours of the MMW dataset~\cite{dv05-ng17-23}. Our DA-based approaches have provided robustness against dynamic factors' variation between training and test data and achieved a 0.898 F1-score.
\end{enumerate}

The study is organized as follows - Section~\ref{sec:literature} discusses recent works on MMW radar.  Data-gathering and processing techniques are described in Section~\ref{sec:dataset}. The model training mechanism is elaborated in Section~\ref{sec:method}. Section~\ref{sec:experiments} reports the experimental settings, results, and findings, and Section~\ref{sec:conclusion} concludes the study.

\section{Related Works}
\label{sec:literature}
In this section, we briefly discuss the MMW radar-based approaches: First, we discuss the literature where MMW radar is deployed as an individual sensor unit and, second, when MMW radar is associated with other sensor units.
\subsection{MMW Radar-based approaches} 
MMW radar is widely used for various research purposes, including human motion behavior detection~\cite{zhang2018real}, periodic heart rate measurements~\cite{zhao2020heart}, navigation~\cite{lu2020see}, tracking multiple persons~\cite{zhao2019mid}, pose detection~\cite{sengupta2020mm} and estimation~\cite{Li2020}, human activity recognition~\cite{singh2019radhar}, security~\cite{dong2021}, autonomous driving~\cite{zhao2020, toker2020mmwave}, human face and emotion recognition~\cite{challa2021face, dang2022emotion}.~\cite{zhang2018real} senses the micro-Doppler information of the user and detects the motion behavior. mBeats~\cite{zhao2020heart} measures heart rate in different poses in an unobtrusive manner.~\cite{Li2020} leverages two radar data points for human pose estimation. Two radar points are used to generate heatmaps, and Convolutional Neural Network is deployed to transform the two-dimensional heatmaps into a human pose.~\cite{singh2019radhar} proposes RadHAR, a low-cost human activity recognition framework that leverages sparse and non-uniform point clouds from an MMW radar.~\cite{dong2021} has proposed a secure method for speaker verification using MMW radar to prevent an adversarial attack and enhance the home security. Authors have utilized the radar to capture both vocal cord vibration and lip motion for identifying speakers.~\cite{zhao2020} proposes a new MMW radar point cloud classification algorithm to improve human-vehicle classification accuracy in complex scenes of autonomous driving.

\subsection{MMW Radar with other sensor modalities}
Aside from being deployed as a standalone sensor modality, MMW radar has also been widely used as a vision-based sensor modality, mainly to overcome the critical limitations of visibility of vision-based approaches in the absence of light.
~\cite{chang2020spatial} has proposed a fusion-based approach where MMW radar is used for obstacle detection and features are fused with the vision sensor feature.~\cite{zhang2019} proposes a new radar-camera fusion system that takes into consideration the error bounds of the two different coordinate systems from the heterogeneous sensors. The authors have utilized a new fusion-extended Kalman filter to adapt to the heterogeneous sensors.~\cite{shuai2021millieye} proposes a framework called milliEye, a lightweight MMW radar and camera fusion system for robust object detection on edge platforms. milliEye possesses several key advantages over existing sensor fusion approaches.~\cite{wang2021rodnet} proposes a deep radar object detection network, named RODNet, which is cross-supervised by a camera-radar fused algorithm. RODNet considers object detection in various driving scenarios.

\subsection{Differences with existing approaches}
Advanced driver assistance systems (ADAS) and autonomous driving applications in the automotive sector utilize MMW radar~\cite{toker2020mmwave} for more user-centric features. For example, MMW radar can identify obstacles like vehicles and pedestrians on the road and notify the driver. Gao et al. have created a large raw radar dataset for various objects using data from Angle of Arrival, Doppler Velocity, and other sources under different scenarios~\cite{gao2019}.~\cite{xu2022object} detects cyclists, pedestrians, and vehicles by converting the position, velocity, and RCS information into images obtained from the MMW radar.~\cite{toker2020mmwave} has utilized range velocity heatmaps for differentiating between vehicles and pedestrians. Moreover,~\cite{kosuge2022mmwave} has proposed mmWave-YOLO for detecting humans, cones, bicycles, signboards, fences, and riders using radar cross-section data. Most of the previous studies have focused on detecting cars, people, and bicycles using MMW radar data. The previous studies have focused on larger subjects, while our study examines smaller objects typically found indoors or outdoors. 

In this study, the statistical features of radial distance, signal-to-noise ratio, and the statistical nature of the noise have been investigated. MMW radar enabling object recognition may be a promising technology for the smart home industry, which has potential uses in energy optimization, security, and human-machine interfaces. Therefore, we are interested in learning more about MMW radar's particular modality's ability to recognize objects. From an implementation point of view, we have considered two scenarios. Firstly, objects of smaller dimensions such as dimes, lead pencils, plastic sheets, wood, and quarters have been investigated when the radar is in a static state. For the dynamic state of MMW radar, we have considered five different objects: Unnamed Ground Vehicle (UGV), water bottles, plastic sheets, paper, and clothes. In a static state, we are thinking of finding objects on the floor or in a green field. While the radar emits signals vertically, it requires additional time to receive the signals. If it is in motion, it will be difficult to receive the signal. For this reason, we position the radar horizontally when it is static. The moving radar emits horizontal signals. At the state of motion, the radar receives the signal in a vertical orientation. That's why we place the radar vertically when the radar is dynamic. According to the position and status of the MMW radar, we want to measure how well it performs. When the position and condition of the MMW radar change, different objects are taken into consideration. Due to the size of the object-to-wavelength ratio and different environmental ambience, detecting small objects by MMW radar involves additional technical challenges such as a high signal-to-noise ratio, obfuscation, etc. 

As we are getting time-series MMW radar data, we have applied a 1D Convolutional Neural Network (CNN) to recognize the objects. Other researchers have used 2D CNN~\cite{xu2022object} instead of 1D CNN, but 1D CNN performs satisfactorily due to the nature of our data. The 1D CNN has achieved 95\% accuracy, which is covered in section~\ref{sec:experiments}.


\section{Dataset}
\label{sec:dataset}

mmWaves are short-wavelength electromagnetic waves that fall in the frequency range of 30-300 \textit{GHz} (1 millimeter to 1 centimeter)~\cite{iovescu2020fundamentals}. This radar is used for precisely measuring the distance of an object, its velocity, and its angle with no or little interference~\cite{iovescu2020fundamentals}. In this section, we describe the data collection procedure and data preprocessing of this systematic study.

\subsection{Data Collection}

\begin{figure}[!ht]
	\centering
	\begin{minipage}{.27\linewidth}
		\centering
		\subfloat[]{\includegraphics[height=3cm,width=\textwidth]{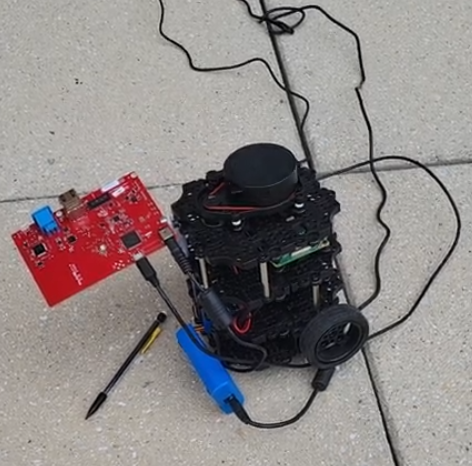}}
        \label{fig:fig(a)}
	\end{minipage}
	\hspace{2em}
	\begin{minipage}{.25\linewidth}
		\centering
		\subfloat[]{\includegraphics[height=3cm,width=\textwidth]{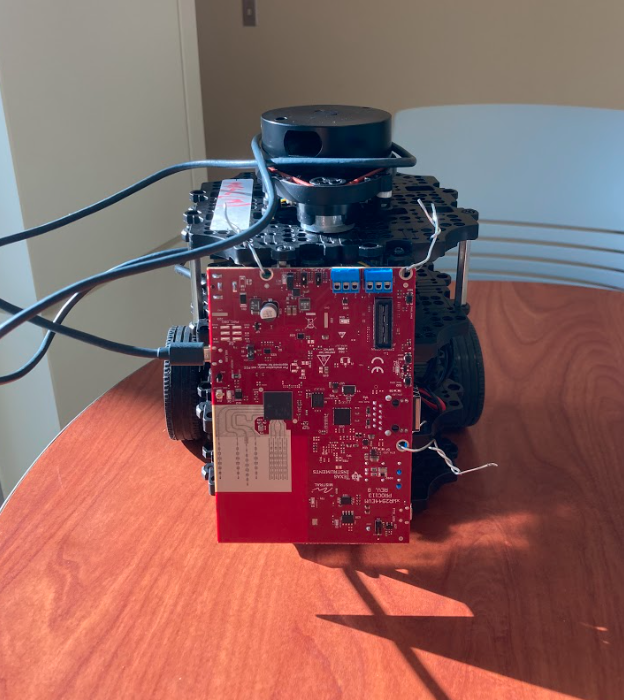}}    
	\end{minipage}
	\hspace{2em}
	\begin{minipage}{.27\linewidth}
		\centering
		\subfloat[]{\includegraphics[height=3cm,width=\textwidth]{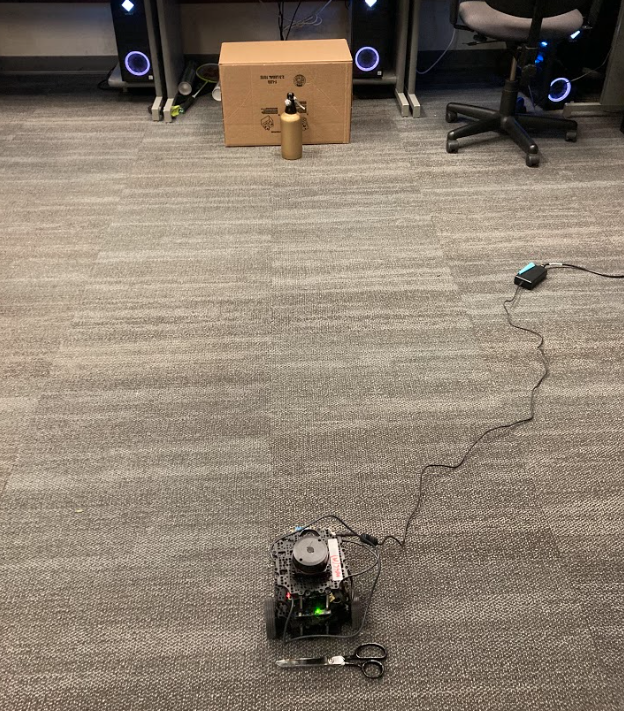}}
	\end{minipage}
	\caption{(a) The red MMW radar is integrated with Turtlebot3, positioned over a black lead pencil, powered by black cables, and charged using a blue charger, (b) integrated with Turtlebot3, the red MMW radar emits horizontal signals to recognize objects, (c) collecting data at a distance of 84 inches in lab light by moving the radar at a constant velocity.}
 
	\label{fig:data_collection_device2}

\end{figure}

Firstly, we consider MMW radar as static. We leverage an MMW radar that is integrated with a commodity, low-cost robot tool kit called TurtleBot3~\footnote{https://emanual.robotis.com/docs/en/platform/turtlebot3/overview/} as depicted in Figure~\ref{fig:data_collection_device2}(a). Secondly, we consider MMW radar as dynamic. Figure~\ref{fig:data_collection_device2}(b) depicts the vertical positioning of the radar for data collection. Communication between TurtleBot3 and MMW radar is interfaced with the TurtleBot3 via the Raspberry Pi 4B. The radar operates with a sampling frequency of 76–81 \textit{GHz} and a wavelength of about 4 mm (approximately 0.16 inches). MMW radar provides a 3-axis radial distance, signal-to-noise ratio (SNR), and noise after blending the transmitted and reflected signals from the object that lies in the radar projection path. 

\begin{figure}[!ht]
  \centering
  \subfloat []{\includegraphics[width=0.32\columnwidth]{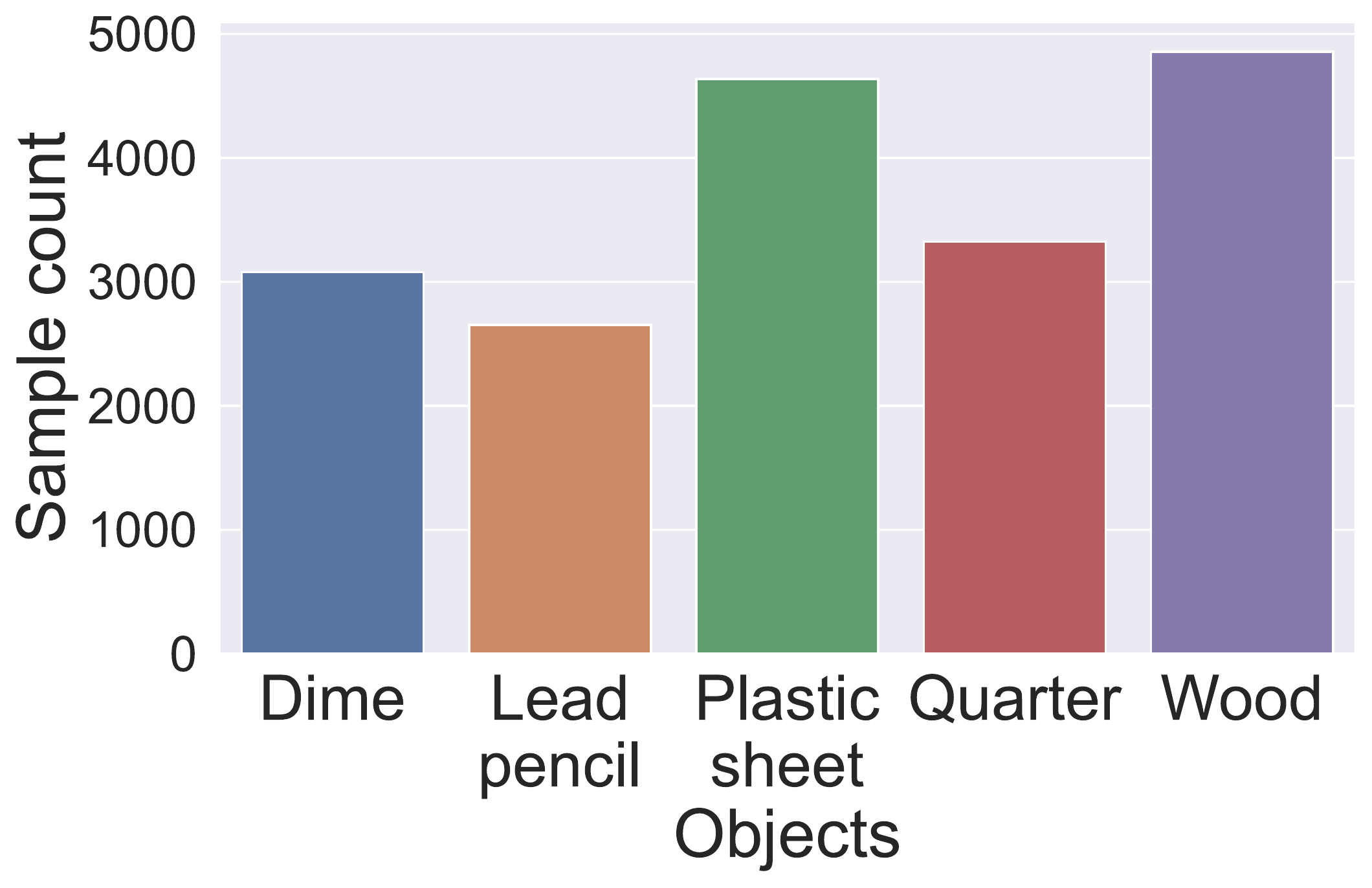}}
  \hfill
  \subfloat []{\includegraphics[width=0.32\columnwidth]{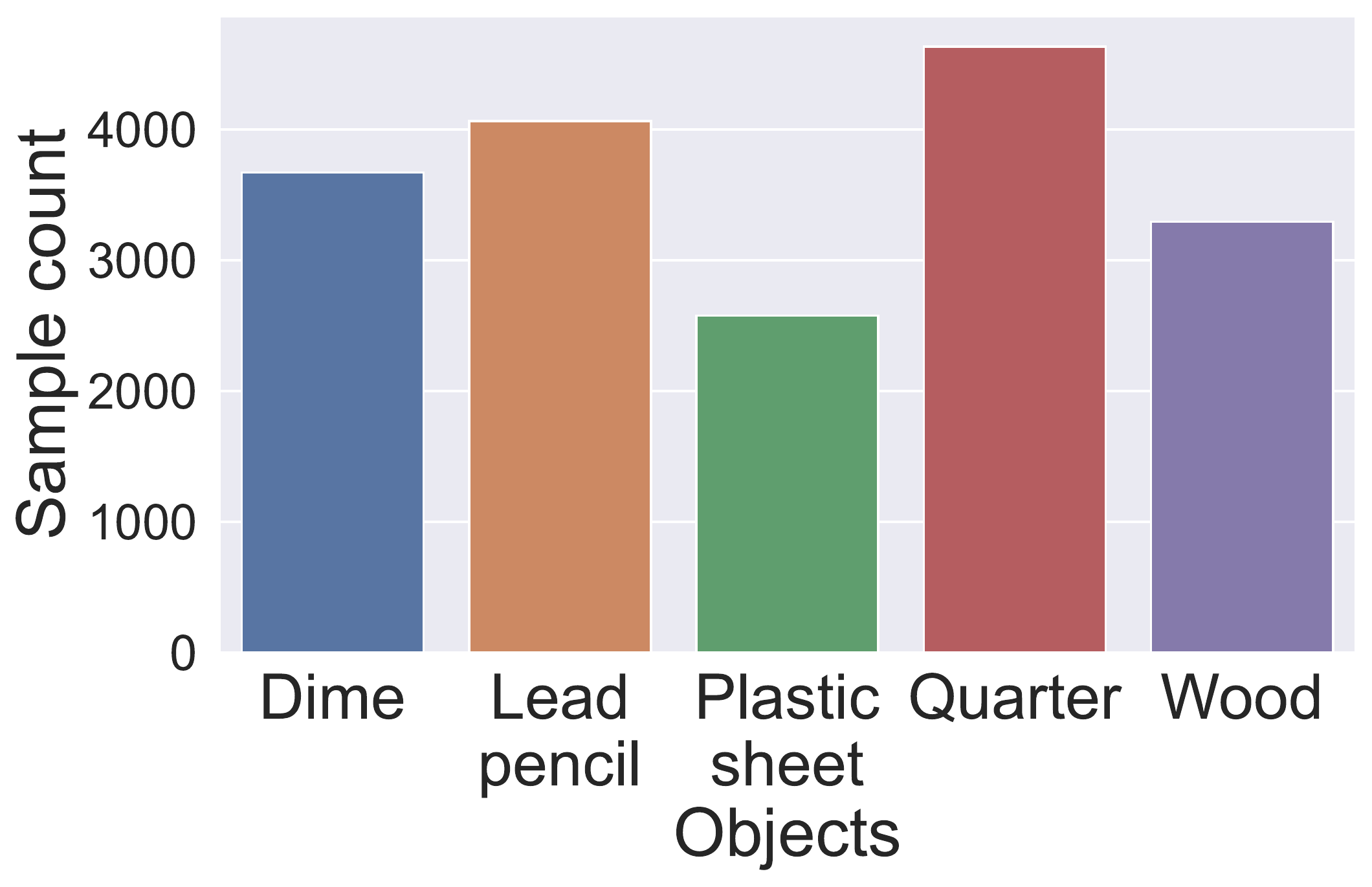}}
  \hfill
  \subfloat []{\includegraphics[width=0.32\columnwidth]{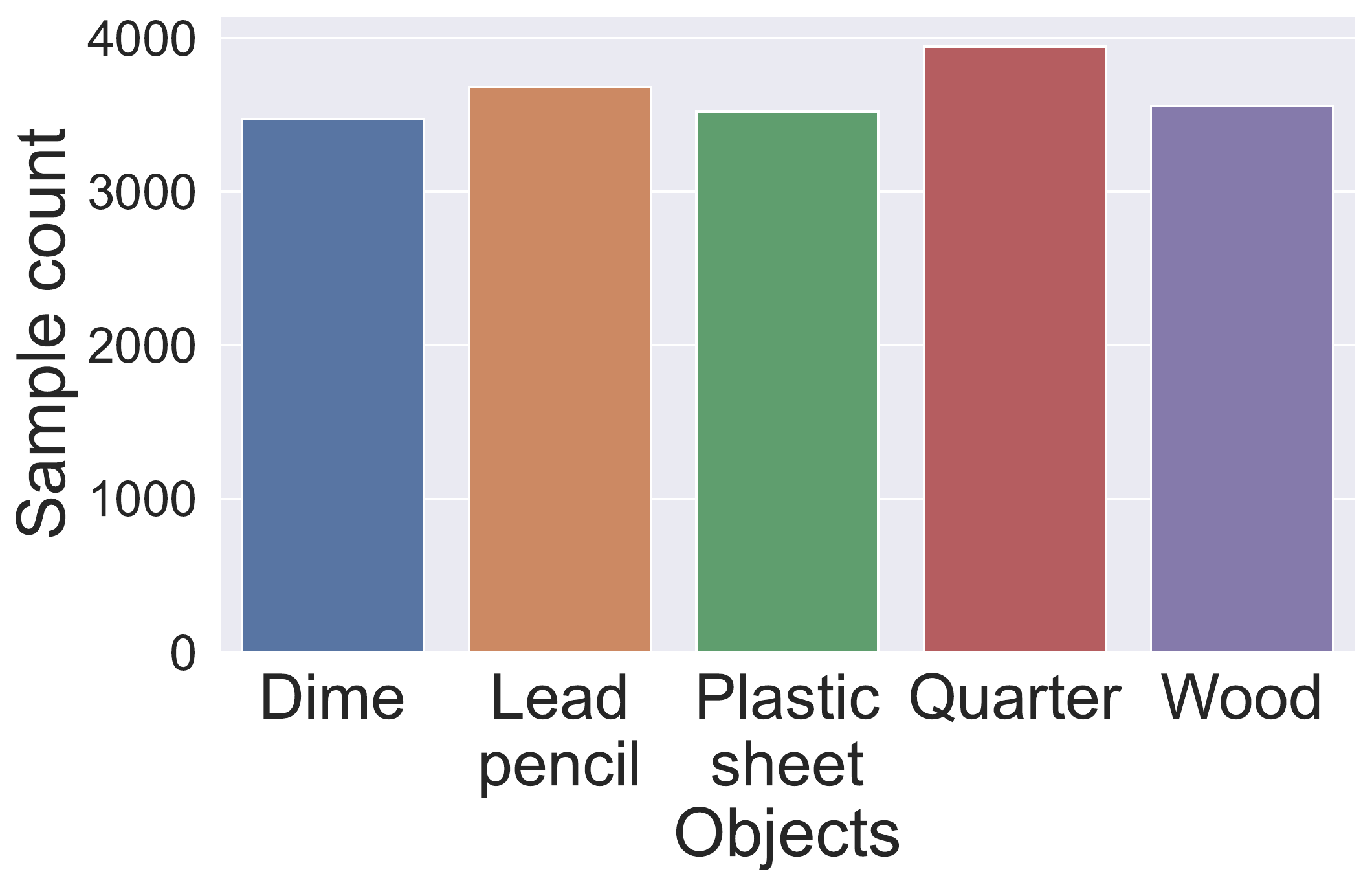}}
  \caption{Data distribution in sunny (L), lablight (M), and night (R) environment for height 7 inches (when the radar is static).}
  \label{fig:data_dis_7}
\vspace{-2ex}
\end{figure}

\begin{figure}[!ht]
  \centering
  \subfloat []{\includegraphics[width=0.32\columnwidth]{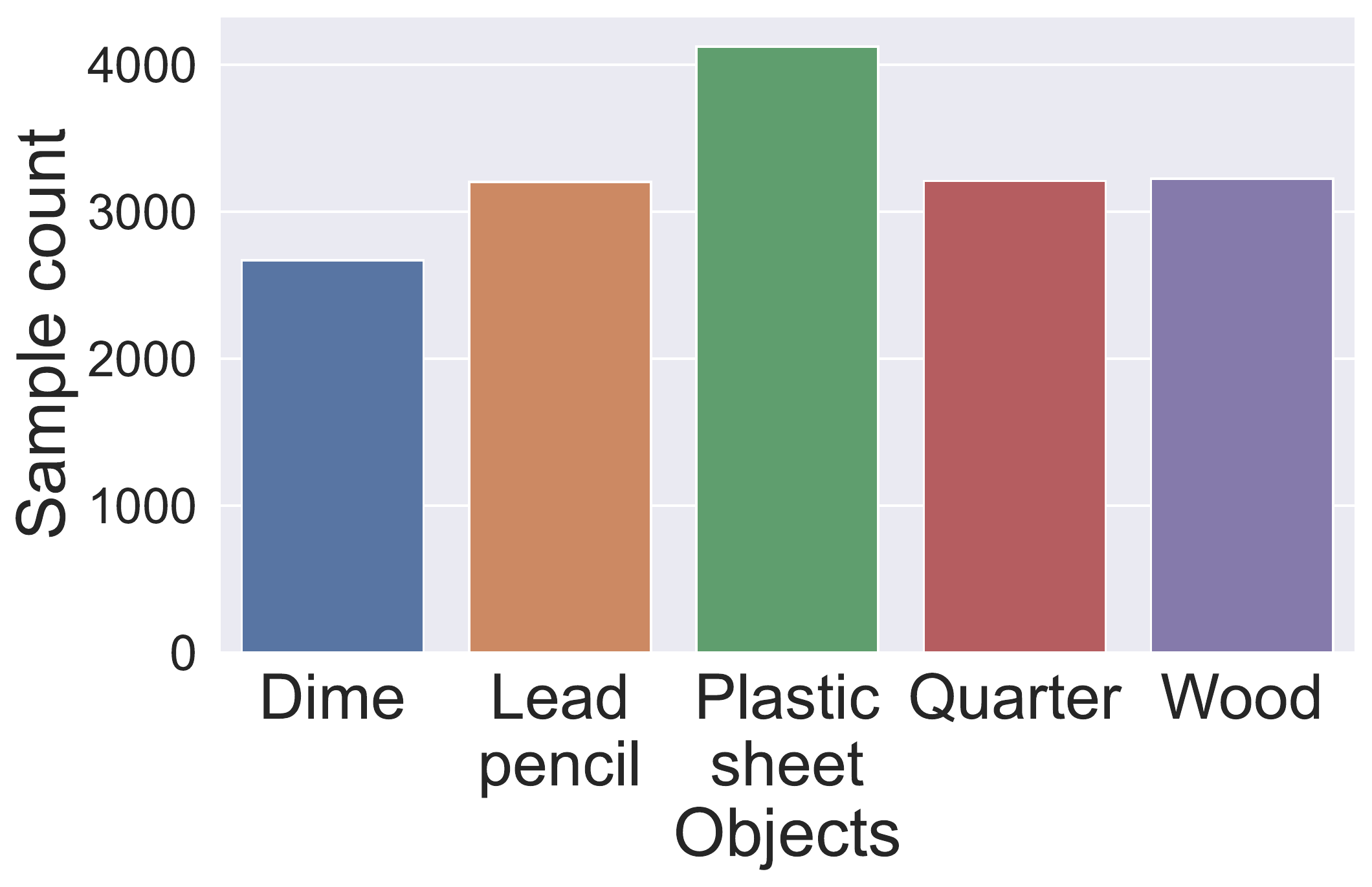}}
  \hfill
  \subfloat []{\includegraphics[width=0.32\columnwidth]{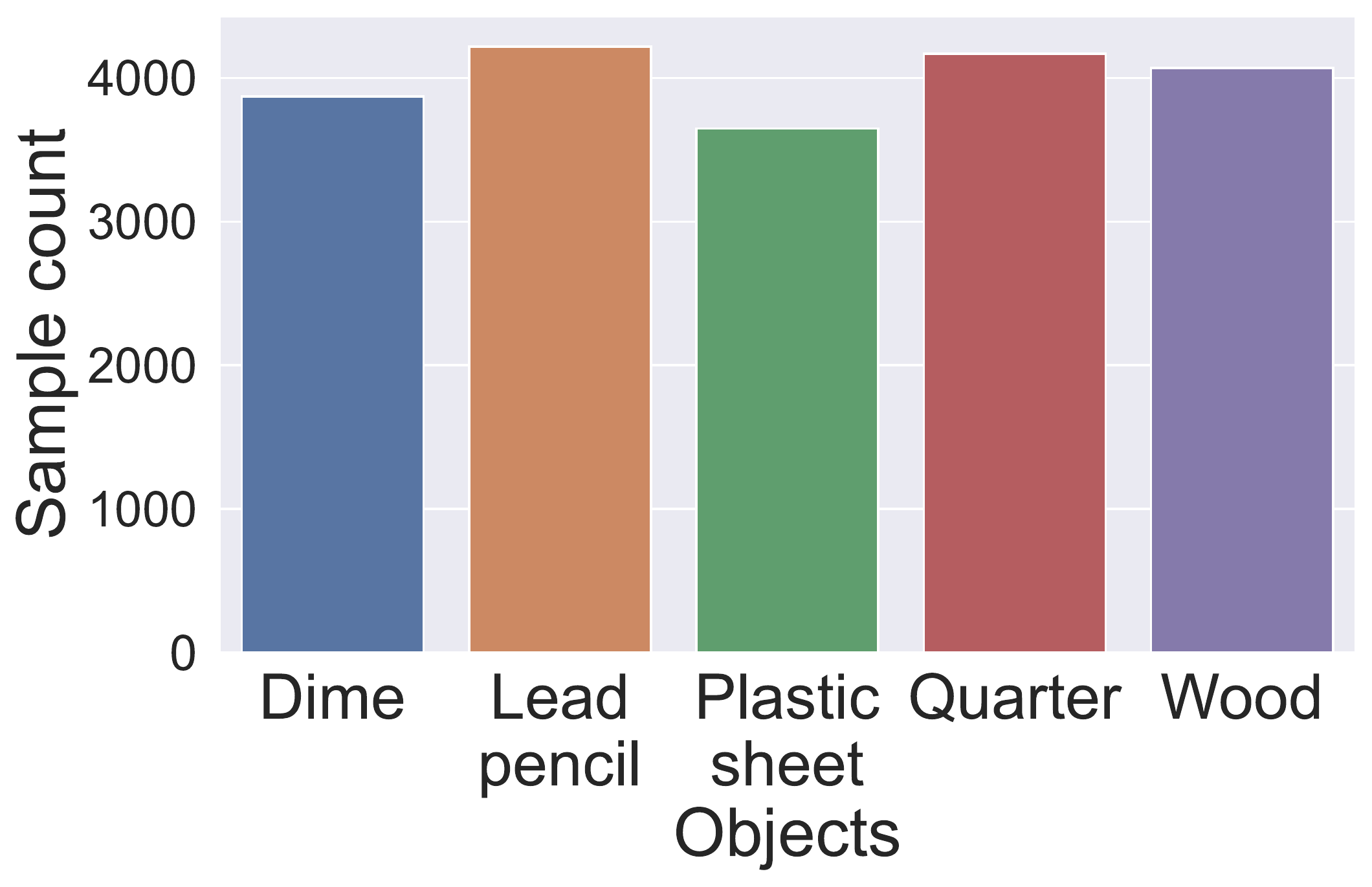}}
  \hfill
  \subfloat []{\includegraphics[width=0.32\columnwidth]{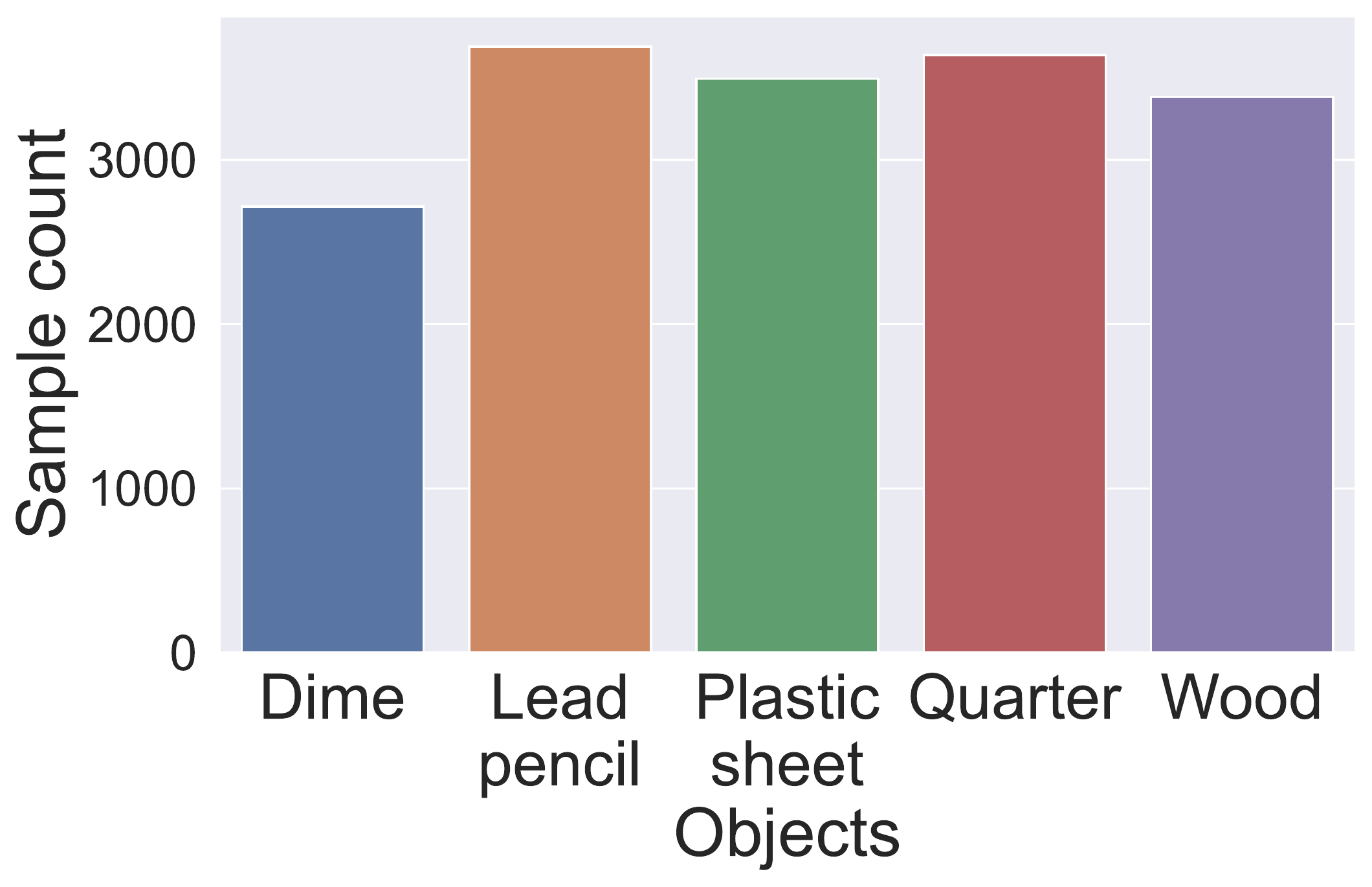}}
  \caption{Data distribution in sunny (L), lablight (M), and night (R) environment for height 53 inches (when the radar is static).}
  \label{fig:data_dis_53}
\vspace{-2ex}
\end{figure}

\begin{figure}[!ht]
  \centering
  \subfloat []{\includegraphics[width=0.32\columnwidth]{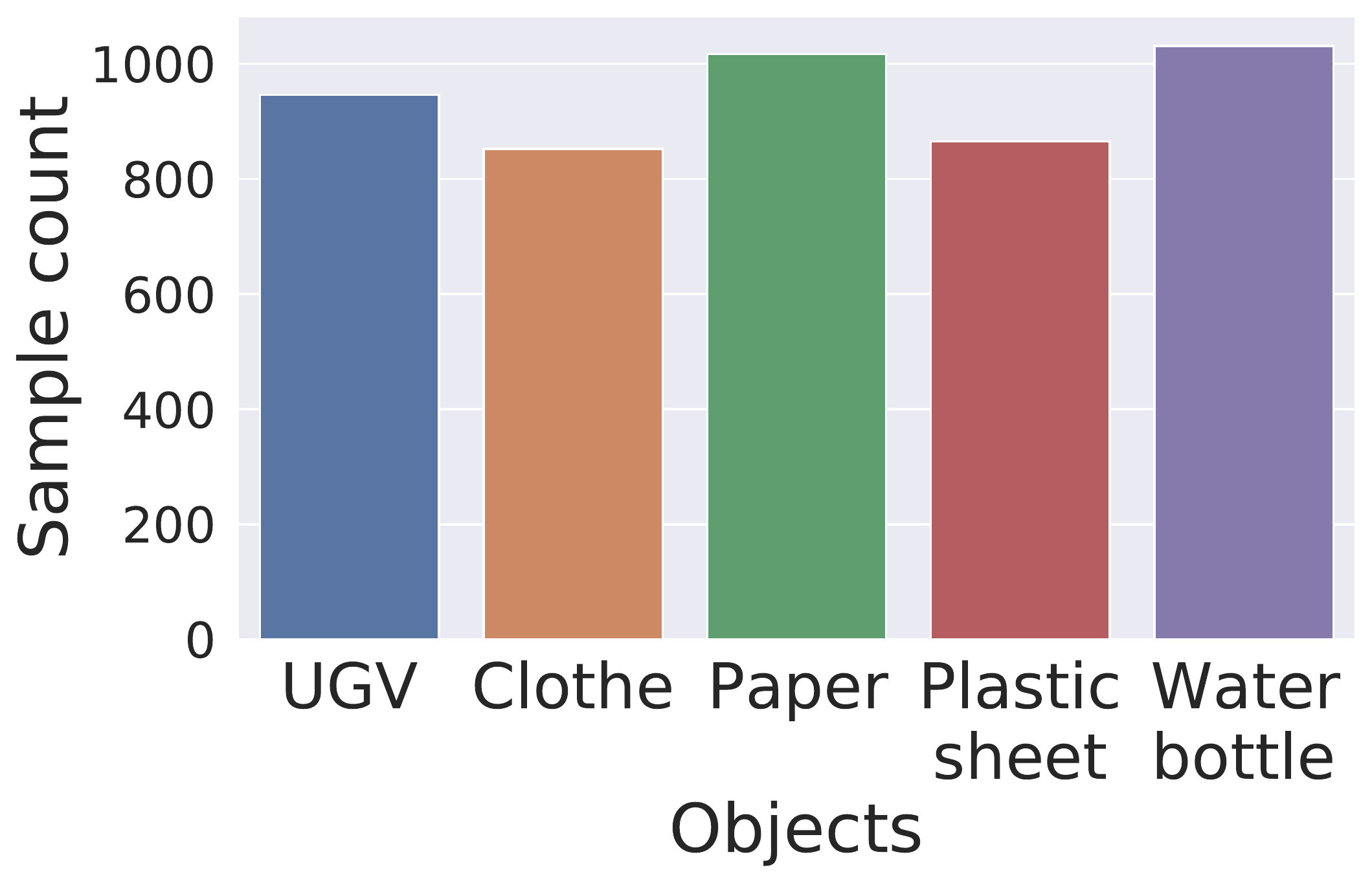}}
  \hfill
  \subfloat []{\includegraphics[width=0.32\columnwidth]{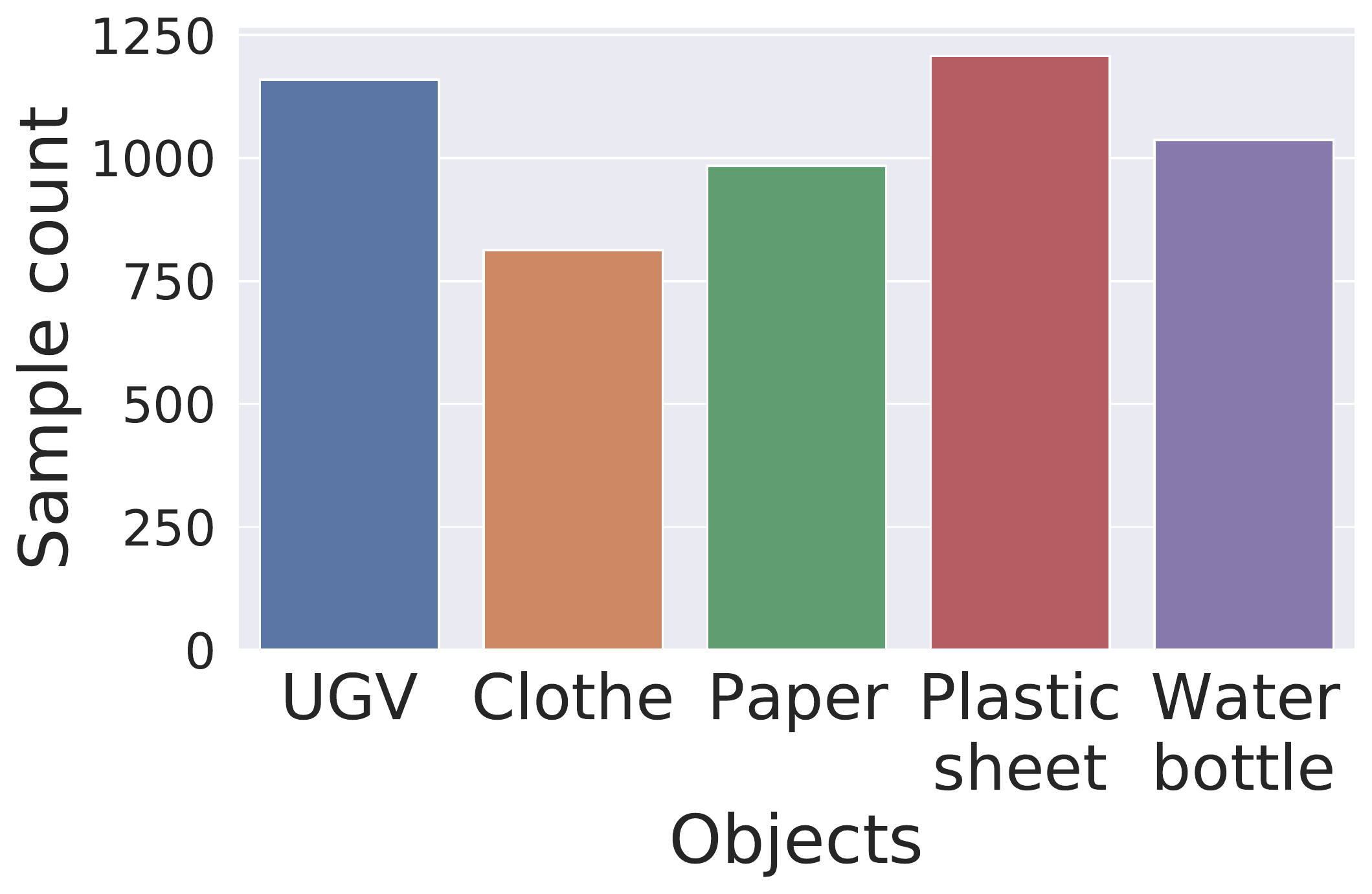}}
  \hfill
  \subfloat []{\includegraphics[width=0.32\columnwidth]{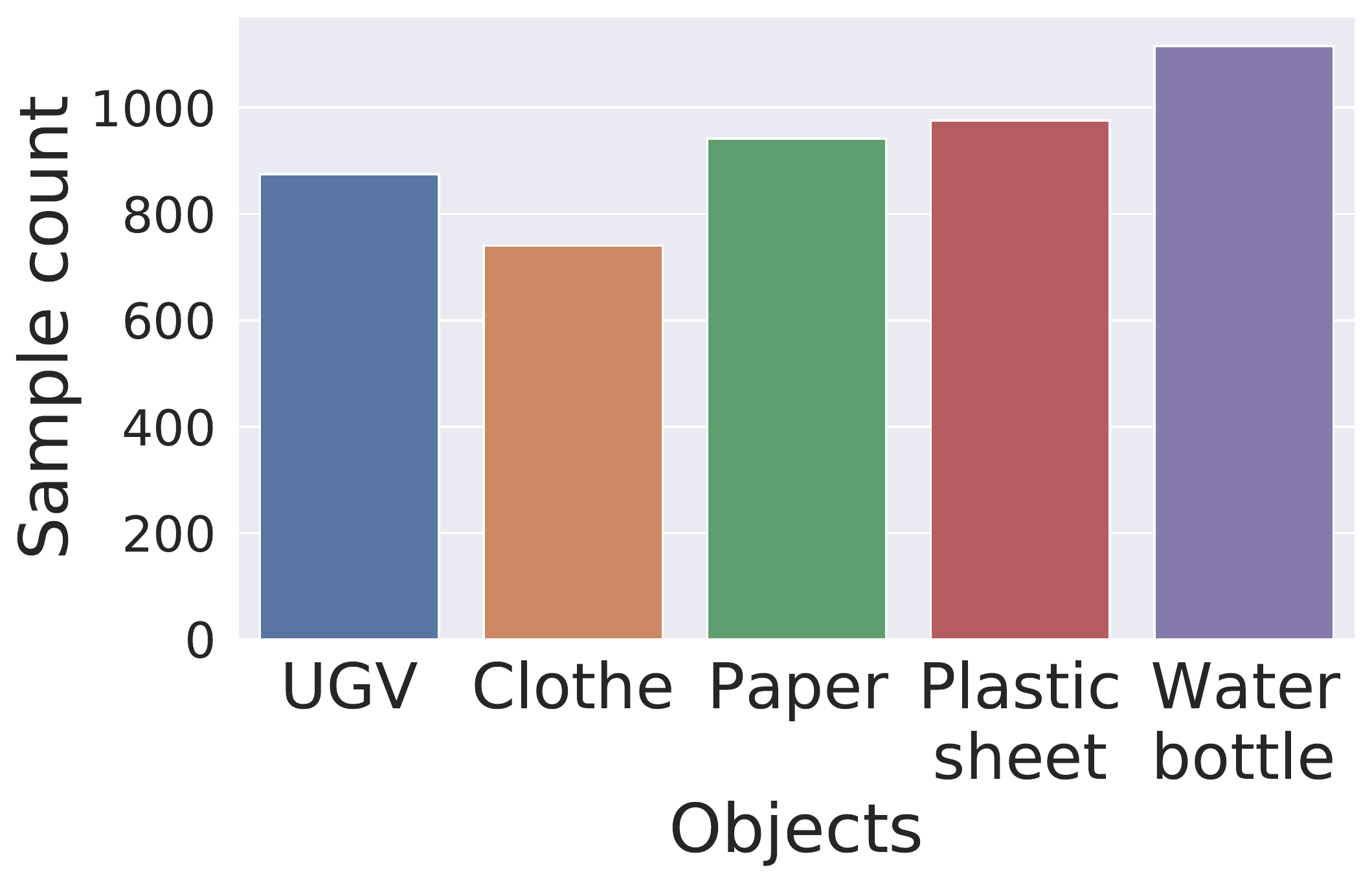}}
  \caption{Data distribution in sunny (L), lablight (M), and night (R) environment for distance 42 inches (when the radar is dynamic).}
  \label{fig:data_dis_42}
\vspace{-2ex}
\end{figure}

\begin{figure}[!ht]
  \centering
  \subfloat []{\includegraphics[width=0.32\columnwidth]{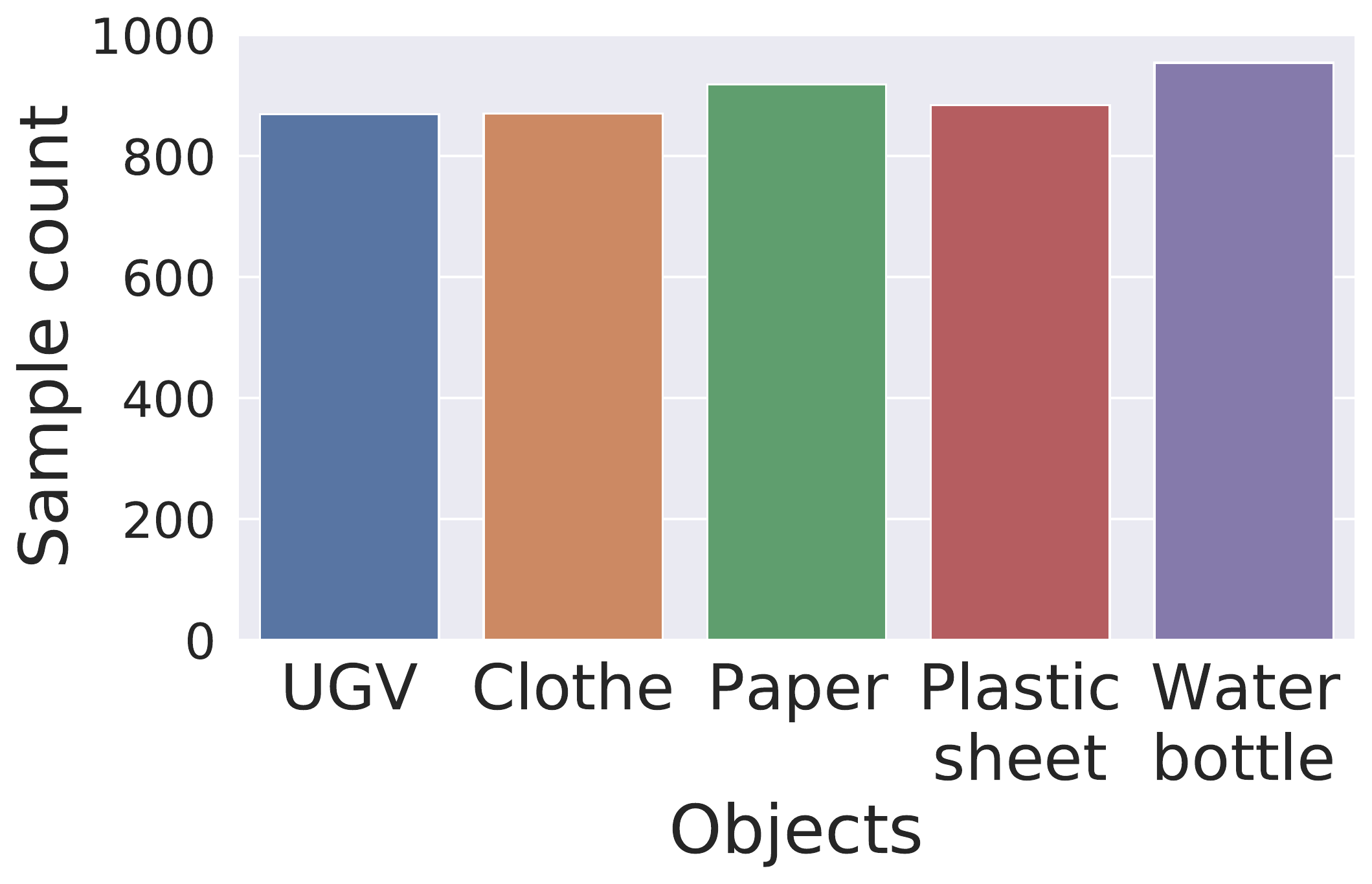}}
  \hfill
  \subfloat []{\includegraphics[width=0.32\columnwidth]{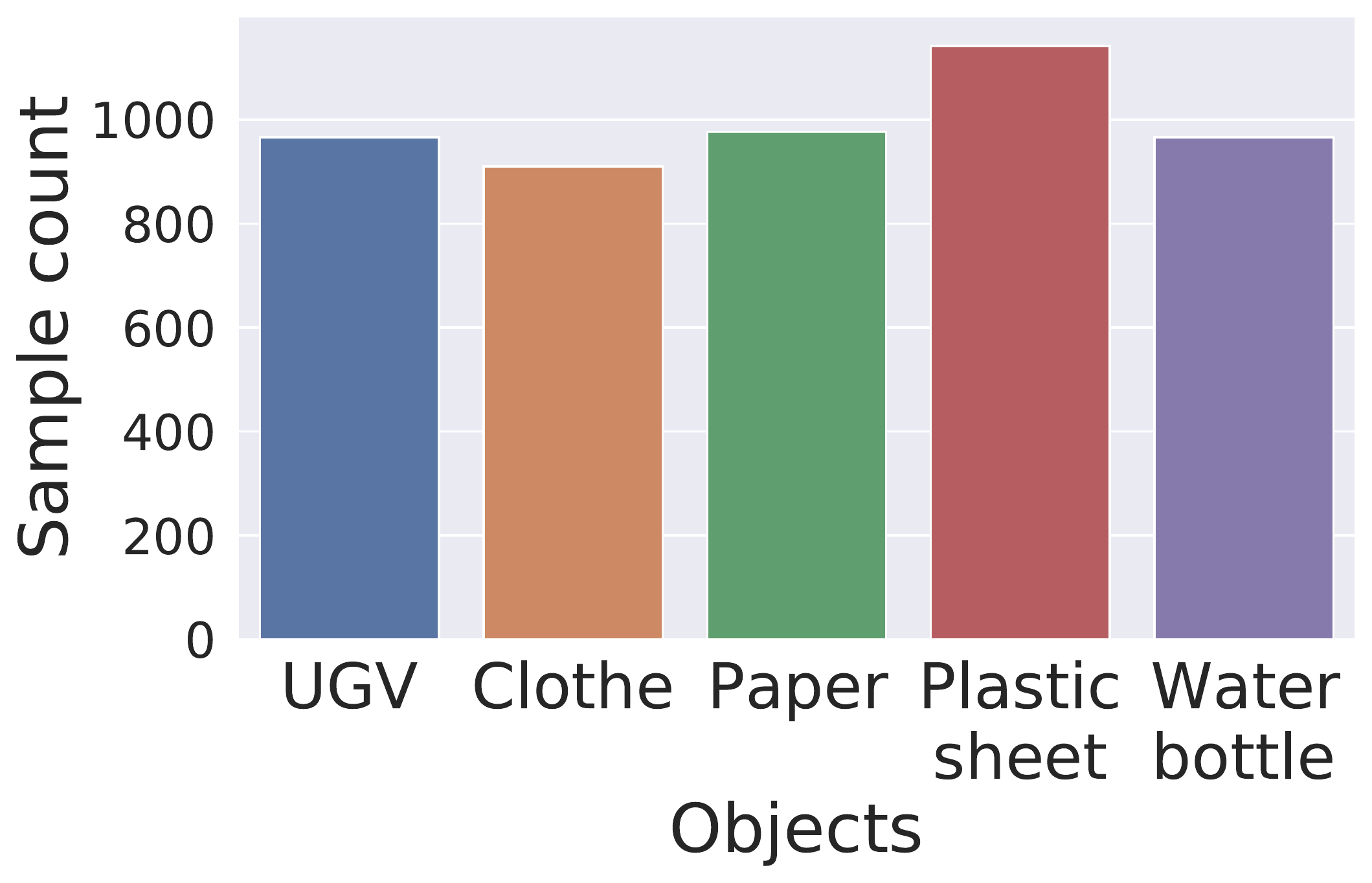}}
  \hfill
  \subfloat []{\includegraphics[width=0.32\columnwidth]{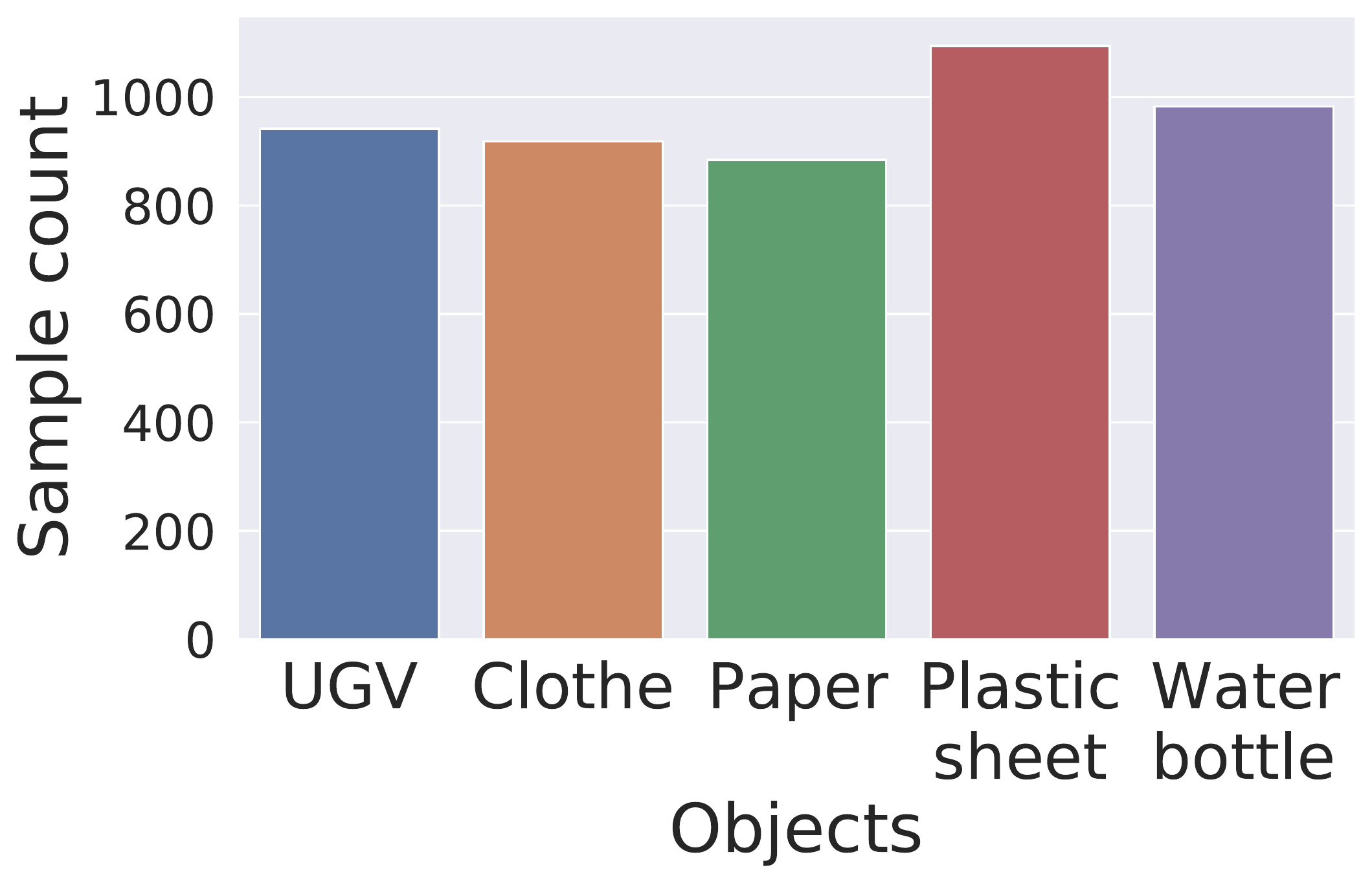}}
  \caption{Data distribution in sunny (L), lablight (M), and night (R) environment for distance 84 inches (when the radar is dynamic).}
  \label{fig:data_dis_84}
\vspace{-2ex}
\end{figure}

In this study, the impact of height and lighting conditions on static radar data collection is examined. Two different heights are considered: placing the TurtleBot3 on the ground, where the radar is 7 inches vertically apart from the ground, and placing it at a height where the radar is 53 inches from the ground. Five different objects ( dime (10 cent US coin), quarter (25 cent US coin), lead pencil, plastic sheet, and wood) made of four different materials are used, and data is collected for two minutes at different angles for each object under three different lighting conditions: sunny, indoor room environment with light (Lablight), and indoor room environment without light (Night). Data collection is done outside with natural light for the sunny condition and inside a lab with white light or without white light for the other two conditions. Figure~\ref{fig:data_dis_7} and~\ref{fig:data_dis_53} depict the data distribution under various lighting conditions for static radar. To study the impact of distance and lighting conditions on moving radar, two different distances are considered: placing the TurtleBot3 on the ground, where the radar is 42 inches horizontally apart from the objects, and placing it in a certain position where the objects are 84 inches away from the radar. Five distinct objects made (UGV, water bottle, plastic sheet, paper, and clothes) of four different materials are used, and data is collected for two minutes at different angles for each object under three different lighting conditions. The radar is moved back and forth towards the objects at different angles with a uniform velocity of 0.025 meters per second, which is shown in Figure~\ref{fig:data_collection_device2}(c). The sample count of objects is lower when the MMW radar is in a dynamic state, as the receiver cannot catch all of the signals. We put five different things either below or in front of the radar, depending on whether the radar sends out signals vertically or horizontally. Overall, six sets of datasets are needed to continue this systematic study. Figure~\ref{fig:data_dis_42} and~\ref{fig:data_dis_84} depict the data distribution under various lighting conditions for moving radar.


\subsection{Data Preprocessing}
Several statistical features such as standard deviation, variation, mean, minimum, and maximum are calculated from the MMW radar data, and the extracted features are further standardized by applying standard scalars from the Sci-kit python library~\cite{scikit-learn}. The normalized dataset is being split in a stratified manner into a 70–30\% ratio, with 70\% of the data used for training and 30\% for testing. This stratification ensures that both training and testing data are representative of each class. The training data is directly fed to the Fully Connected Layer (FCL), regardless of whether the radar is in a static or dynamic state. For recognizing objects with dynamic or static radar at longer distances or heights, we have applied 1D CNN for better recognition. For generating the input for 1D CNN, we have combined 10 samples to create a $40\times16$ frame. Later, flatten the $40\times16$ frame to feed into the 1D CNN layer. To create the $40\times16$ frame, we have first taken samples from the datasets spanning rows 1 through 40 and then rows 2 through 41, which is true for the rest of the data samples. Here, 40 rows indicate 4s of data captured by MMW radar. To ensure consistency, we apply windowing to process the continuous data obtained from the same objects. Each window consists of 40 rows, which corresponds to a 4-second MMW radar data acquisition. Our analysis shows that 4 seconds of data are necessary for accurate recognition.




In the next section, we describe the deep models that are used for our feasibility studies. 
\section{Methodology}

\label{sec:method}
To explore the feasibility of using low-resolution and sparse point cloud data for object recognition, we employ two distinct deep learning frameworks: one that utilizes a FCL, and the other that is based on a CNN. Next, we have used the domain adaptation methods~\cite{ganin2015unsupervised} to increase the robustness of the CNN-based deep learning model. We have described the results of FCL and CNN in section~\ref{sec:experiments}.

\subsubsection{Training Mechanism of Fully Connected Layer}
\begin{figure}[!ht]
    \centering
     \includegraphics[width =\columnwidth]{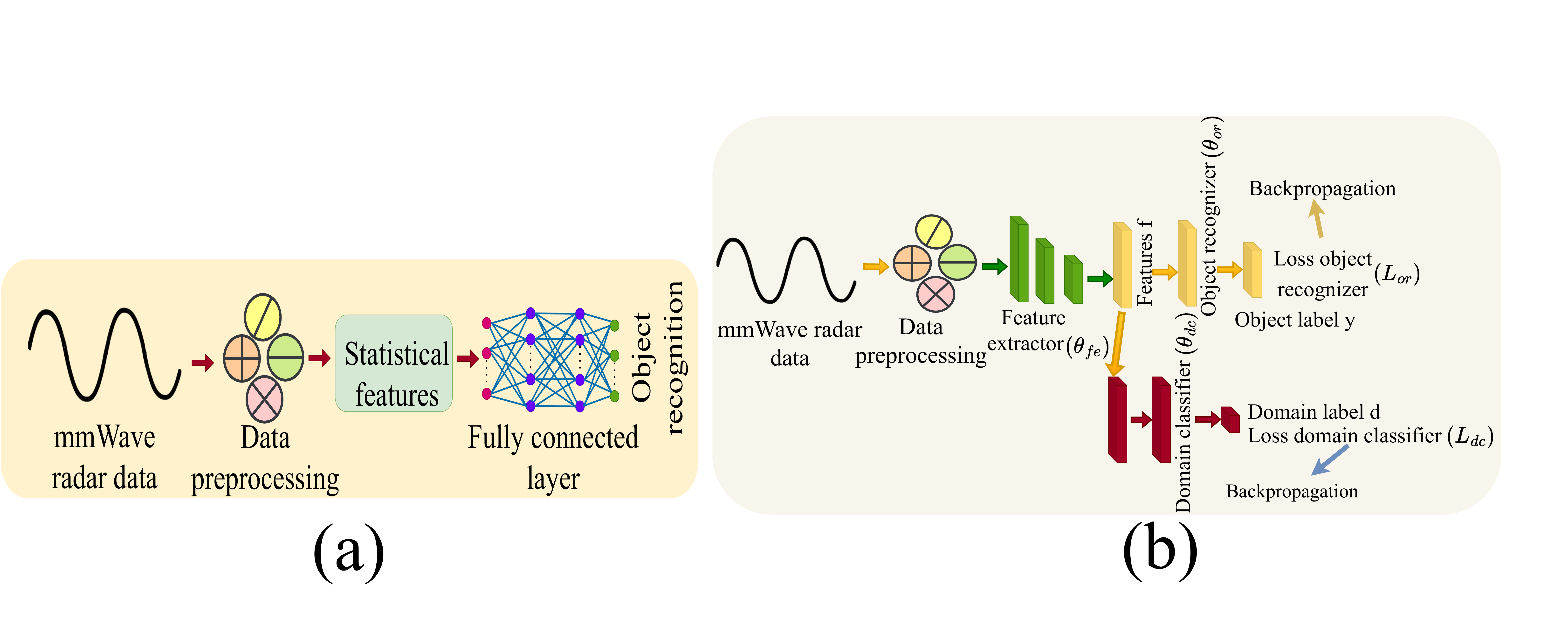}
    \caption{Overall process of object recognition using MMW radar (a) processed MMW radar's data is fed into the FCL for object recognition, (b) processed MMW radar's data is fed into the CNN, which is used for feature extraction, and a FCL is used for recognizing the objects.}
    \label{fig:overall_system}
\end{figure}

 A $3$-layer fully connected Multi-Layer Perceptron (MLP) as the object recognizer train the model in a supervised manner which is illustrated in Figure~\ref{fig:overall_system}(a). The first two layers consist of $16$ computation units followed by a Rectified Linear Unit (ReLU), and the final layer is followed by a Softmax layer. We input the extracted features into the model and train it using the categorical cross-entropy loss in a supervised fashion as described in Equation~\ref{equation:categorical}. 

 \begin{equation}
    \label{equation:categorical}
  L_{cce} = -\sum_{i=1}^C t_i log(p_i)
\end{equation}

where $C$ is the total number of classes (objects), $t_i$ is the ground truth, and $p_i$ is the Softmax probability of the object recognizer.
\vspace{-2ex}
 
 \subsubsection{Training Mechanism of CNN-based Deep Learning Model}
 The CNN-based deep learning framework is illustrated in Figure~\ref{fig:overall_system}(b) and can be categorized into three distinct modules: the feature extraction module (shown in green color), the object recognition module (shown in yellow color), and the domain discriminator module (shown in maroon color). The feature extraction module consists of three distinct 1D Convolutional (Conv1D) layers with ReLU activation layers. On the other hand, the object recognizer module comprises two FC Layers. Finally, the domain adaptation module comprises one FC Layer along with two outputs in the final layer. We have provided details of the architecture settings in Table~\ref{table:cnn1d} for easy reference. 

\begin{table}[!h]
\centering
\caption{Architecture hyper-parameters.}
\label{table:cnn1d}

\begin{tabular}{@{}lllll@{}}
\hline 
Hyper-parameters &  &  & Values &  \\

\hline

Convolution filter no. &  &  & 1x311,1x146,1x64 &  \\
Convolution filter dimension &  &  & 1x20,1x20,1x20 &  \\
Stride &  &  & 2 &  \\
No. of units in object recognizer module &  &  & 64,32,16,5 & \\
No. of units in domain adaptation module &  &  & 64,32,2& \\
\hline
\end{tabular}
\end{table}

The framework has demonstrated promising results in recognizing different objects, and the domain adaptation mechanism has helped to improve the adaptability of the deep learning model. For our feasibility study, we have used the feature extractor and object recognizer modules. We have trained the network in a supervised manner using a labeled dataset that has been collected. A batch-wise categorical cross-entropy loss function has been used, which is described in Equation~\ref{equation:categorical}.

Next, we extend the deep learning model's robustness for object recognition using domain adversarial methods~\cite{ganin2015unsupervised}. We use the same CNN model for recognizing the objects and add a domain classifier (shown in maroon color in Figure~\ref{fig:overall_system}(b)). The domain classifier and feature extractor in our investigation both use an adversarial strategy. Both modules accept features from datasets with and without labels. The domain classifier attempts to predict the source originality of the incoming features, whereas the feature extractor operates to negate this goal. The target dataset is an open dataset without available labels, whereas the source dataset has labels. While both modules process samples of labeled and unlabeled data during adaptation, the object recognizer only uses the labeled dataset from the source domain. When the model has been worked in a semi-supervised manner, fewer labels from the target domain have been used. We use the gradient reversal layer implementation from the work~\cite{ganin2015unsupervised}, as indicated in Equations~\ref{equation:feature_extractor},~\ref{equation:object_recognizer} and~\ref{equation:domain_classifier} to accomplish this strategy. 

\begin{equation}
    \label{equation:feature_extractor}
  \theta _{fe} = \theta _{fe} - \mu \left ( \frac{\partial L^{i}_{or}}{\partial \theta _{fe}}  - \frac{\partial L^{i}_{dc}}{\partial \theta _{fe}} \right )
\vspace{-1ex}
\end{equation}
where $\frac{\partial L_{or}}{\partial \theta_{fe}}$ is the gradient of the loss function object recognizer $L_{or}$ with respect to the feature extractor $\theta _{fe}$, and $\frac{\partial L_{dc}}{\partial \theta_{fe}}$ is the gradient of the loss function domain classifier $L_{dc}$ with respect to the feature extractor $\theta _{fe}$ and $\mu$ is the learning rate.
\vspace{-1ex}
\begin{equation}
    \label{equation:object_recognizer}
  \theta _{or} = \theta _{or} - \mu \frac{\partial L^{i}_{or}}{\partial \theta _{or}}
\vspace{-1ex}
\end{equation}
where $\frac{\partial L_{or}}{\partial \theta_{or}}$ is the gradient of the loss function $L_{or}$ with respect to the object recognizer $\theta_{or}$.
\begin{equation}
    \label{equation:domain_classifier}
  \theta _{dc} = \theta _{dc} - \mu \frac{\partial L^{i}_{dc}}{\partial \theta _{dc}}  
\vspace{-1ex}
\end{equation}
where $\frac{\partial L_{dc}}{\partial \theta_{dc}}$ is the gradient of the loss function $L_{dc}$ with respect to the domain classifier $\theta_{dc}$. 

\section{Systematic Study}
\label{sec:experiments}
  We report our experimental findings in this section. We have executed the experiments on an NVIDIA GeForce RTX 3060 graphics card. In different experiments throughout this systematic study, we consider micro-F1 as the evaluation metric as the data distribution is imbalanced~\cite{hamad2020efficacy}. Micro-F1 score is not biased to the majority samples.

\subsection{When MMW radar is static}
\subsubsection{Impact of Ambience}
In this study, we investigate how different surface materials affect the reflected radar signal and ultimately the recognition of objects. The radar projects a signal from a high elevation onto the surface and receives the reflected signal from various objects in its path. As the surface can be made up of different materials, the reflectivity of the signal can be altered. We conduct experiments to recognize objects in three different environments while maintaining a fixed distance. We collect training and testing samples from the same environment at a fixed height. The findings are represented in Figure~\ref{fig:same_height_cross_environment_7_53}(a) and Figure~\ref{fig:same_height_cross_environment_7_53}(b). For 7 inches and 53 inches of the MMW radar, we have used the FCL and CNN. The recognition performance of FCL and CNN is approximately 85\%.


\subsubsection*{Key Takeaways}
\begin{figure}[!ht]

\subfloat[]{%
  \includegraphics[width=\columnwidth]{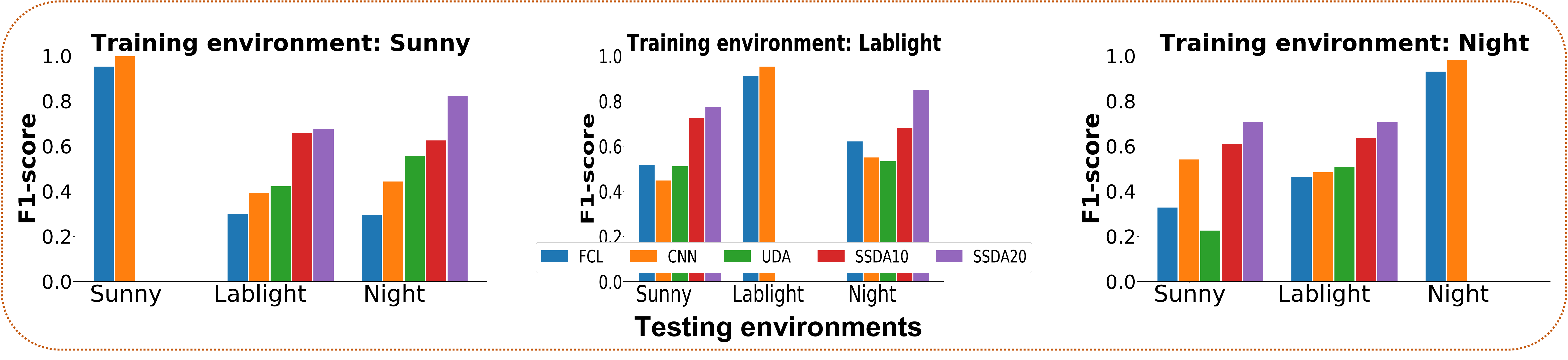}%
}

\subfloat[]{%
  \includegraphics[width=\columnwidth]{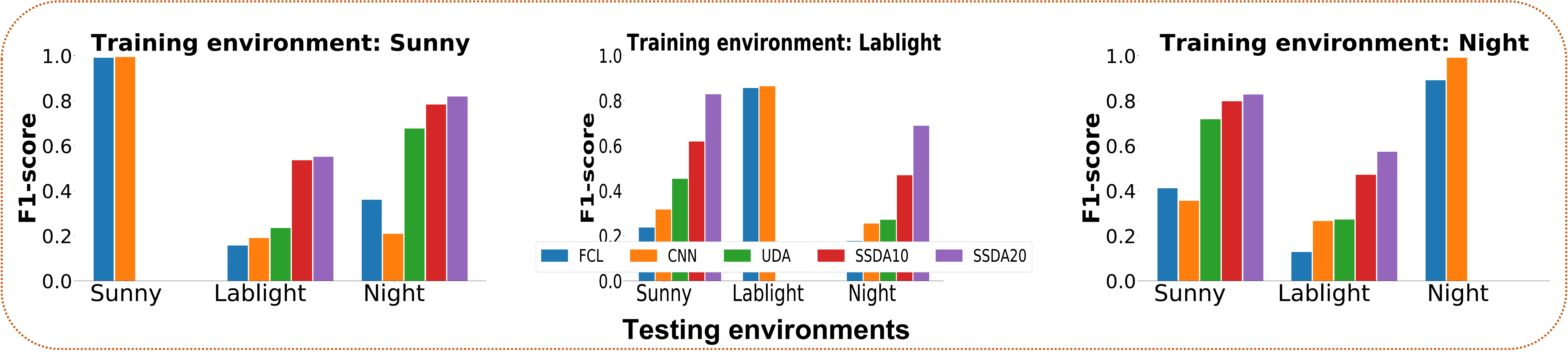}%
}

\caption{Performance of different deep learning models under two conditions: same ambiance and cross-ambiance, for similar heights (a) for a vertical height of 7 inches, (b) for a vertical height of 53 inches.}

\label{fig:same_height_cross_environment_7_53}

\end{figure}

\begin{itemize}
    \item A sunny environment (outdoor) provides better recognition performance than an indoor environment (lab environment with light and without light). It's true both for FCL and CNN methods. Here, the outdoor environment is made of a concrete surface, whereas the indoor environment's surface is an office floor carpet that is typically made of nylon, olefin, or polypropylene.    
    
    \item Within an indoor environment, FCL's performance drops even more when the height is increased, but CNN can hold the performance due to its complex feature-learning nature. We hypothesize that as the height increases, the projected radar signal covers more area compared to the lower height. As the area coverage increases, the reflected signal is impacted by the surrounding objects in the lab that are not recognized by FCL. CNN can recover it using more complex features, which are shown in Figure~\ref{fig:same_height_cross_environment_7_53}(b).
    
\end{itemize}


\begin{figure}[!ht]

\subfloat[]{%
  \includegraphics[width=\columnwidth]{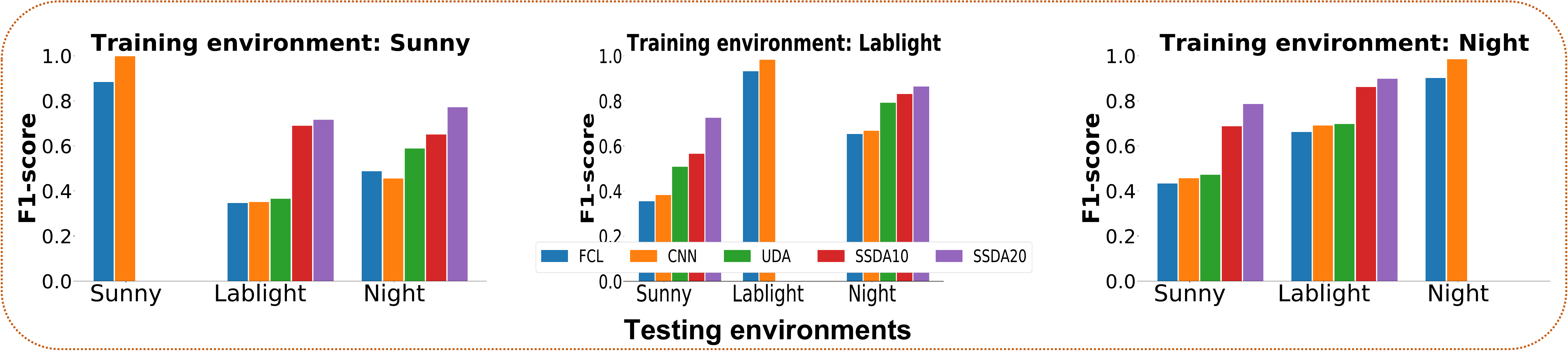}%
}

\subfloat[]{%
  \includegraphics[width=\columnwidth]{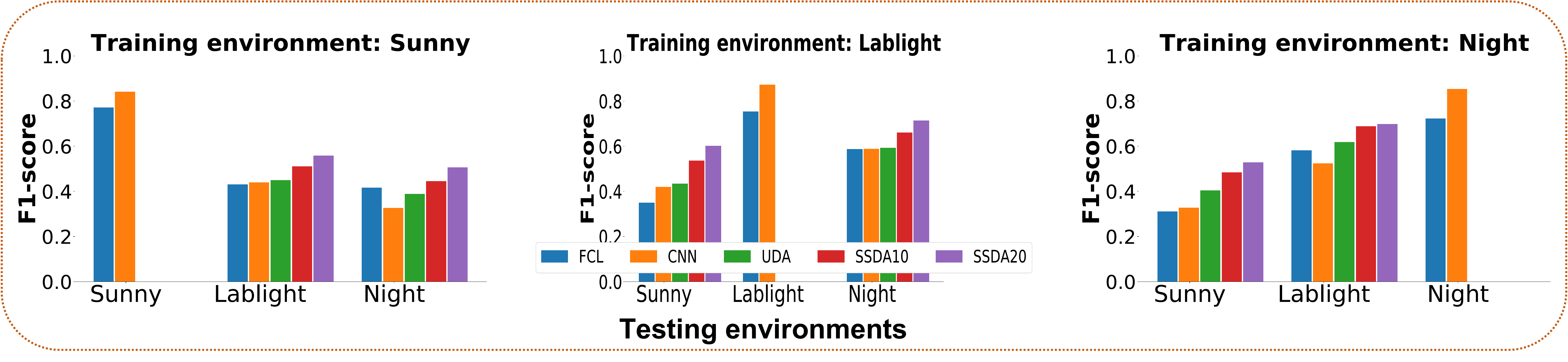}%
}

\caption{Performance of different deep learning models under two conditions: same and cross-ambiance for similar distances. (a) For a horizontal distance of 42 inches, (b) for a horizontal distance of 84 inches.}

\label{fig:same_distance_cross_environment_42_84}

\end{figure}

\subsubsection{Impact of Cross-Ambience}
In this experiment, we study the impact of the collected data in different environments. Similar to the previous study, during training and testing, we leverage the dataset from the same height but vary in respect of the collected environment. Consider two datasets of 53 inches in height—one from the sunny environment and another one from the lablight environment. We train the model using the sunny environment dataset and test it in the lablight environment. This study's experimental findings are summarized in Figure~\ref{fig:same_height_cross_environment_7_53}(a) and Figure~\ref{fig:same_height_cross_environment_7_53}(b).
\subsubsection*{Key Takeaways}
\begin{itemize} 
\item When the ambience of the training and testing datasets differs, the performance drops significantly, and it drops even more when the height is increased. Figure~\ref{fig:same_height_cross_environment_7_53}(a) and Figure~\ref{fig:same_height_cross_environment_7_53}(b) indicate that under cross-ambience evaluation, height 53 performs even worse than height 7. We have used an unsupervised domain adaptation (UDA) and semi-supervised domain adaptation (SSDA) method with 10\% and 20\% labeled data from the target domain to solve this problem. Figure~\ref{fig:same_height_cross_environment_7_53}(a) and Figure~\ref{fig:same_height_cross_environment_7_53}(b) demonstrate that SSDA significantly improves performance while UDA improves performance by less amount. The F1-score rises from 0.443 to 0.822 after utilizing SSDA (shown in Figure~\ref{fig:same_height_cross_environment_7_53}(a)) with training samples from a sunny environment and testing samples from a night environment. It is applicable to other cross-ambience situations as well.
\end{itemize}

\subsubsection{Impact of Cross-Height}
In the cross-ambience study, we find that the performance of the model is severely affected due to the heterogeneity of the data distribution caused by different ambiences. In this study, we aim to investigate the performance of the model when using similar ambience data samples that have been collected from different heights for training and testing purposes.
Table~\ref{tab:same_environment_cross_height} represents the experimental findings.
\subsubsection*{Key Takeaways}
\begin{itemize} 
\item Similar to cross-ambience, cross-height also yields sub-optimal results. We observe a similar pattern for the cross-height and cross-ambience settings used in model training and testing. Table~\ref{tab:same_environment_cross_height} demonstrates that SSDA significantly
improves performance while UDA improves performance by less amount. The F1-score rises from 0.238 to
0.834 (SSDA 10\%) and 0.878 (SSDA 20\%) (shown in Table~\ref{tab:same_environment_cross_height}) with training and testing samples from
the sunny environment but at different heights. It is applicable to other cases also. 
\end{itemize}

\begin{table}[htbp]
\centering
\caption{Performance of different deep learning models
under the same ambience, for
different heights.}
\label{tab:same_environment_cross_height}
\resizebox{.95\linewidth}{!}{%
\begin{tabular}{@{}|c|c|c|c|c|c|c|@{}}
\hline
    
    \begin{tabular}[c]{@{}c@{}}Training\\ environment \\ \end{tabular} &
    \begin{tabular}[c]{@{}c@{}}Testing\\ environment \\ \end{tabular} &
    \begin{tabular}{c}F1-score (FCL)\end{tabular} &
    \begin{tabular}{c}F1-score (CNN)\end{tabular} &
    \begin{tabular}{c}F1-score \\(UDA)\end{tabular} &
    \begin{tabular}{c}F1-score \\(SSDA 10\%)\end{tabular} &
    \begin{tabular}{c}F1-score \\(SSDA 20\%)\end{tabular} \\ 
    \hline

  \begin{tabular}[c]{@{}c@{}}Sunny(53)\end{tabular} &
  \begin{tabular}[c]{@{}c@{}}Sunny(7)\end{tabular} &
  \begin{tabular}[c]{@{}c@{}}0.158\end{tabular} &
  \begin{tabular}[c]{@{}c@{}}0.238\end{tabular} &
  \begin{tabular}[c]{@{}c@{}}0.312\end{tabular} &
  \begin{tabular}[c]{@{}c@{}}0.834\end{tabular} &
  \begin{tabular}[c]{@{}c@{}}0.878\end{tabular} \\
  \hline

  \begin{tabular}[c]{@{}c@{}}Night(53)\end{tabular} &
  \begin{tabular}[c]{@{}c@{}}Night(7)\end{tabular} &
  \begin{tabular}[c]{@{}c@{}}0.346\end{tabular} &
  \begin{tabular}[c]{@{}c@{}}0.217\end{tabular} &
  \begin{tabular}[c]{@{}c@{}}0.250\end{tabular} &
  \begin{tabular}[c]{@{}c@{}}0.531\end{tabular} &
  \begin{tabular}[c]{@{}c@{}}0.799\end{tabular} \\
  \hline

  \begin{tabular}[c]{@{}c@{}}Lablight(53)\end{tabular} &
  \begin{tabular}[c]{@{}c@{}}Lablight(7)\end{tabular} &
  \begin{tabular}[c]{@{}c@{}}0.262\end{tabular} &
  \begin{tabular}[c]{@{}c@{}}0.329\end{tabular} &
  \begin{tabular}[c]{@{}c@{}}0.371\end{tabular} &
  \begin{tabular}[c]{@{}c@{}}0.495\end{tabular} &
  \begin{tabular}[c]{@{}c@{}}0.587\end{tabular} \\

 \hline

  \begin{tabular}[c]{@{}c@{}}Sunny(7)\end{tabular} &
  \begin{tabular}[c]{@{}c@{}}Sunny(53)\end{tabular} &
  \begin{tabular}[c]{@{}c@{}}0.107\end{tabular} &
  \begin{tabular}[c]{@{}c@{}}0.193\end{tabular}&
  \begin{tabular}[c]{@{}c@{}}0.342\end{tabular} &
  \begin{tabular}[c]{@{}c@{}}0.645\end{tabular} &
  \begin{tabular}[c]{@{}c@{}}0.909\end{tabular} \\
  \hline

  \begin{tabular}[c]{@{}c@{}}Night(7)\end{tabular} &
  \begin{tabular}[c]{@{}c@{}}Night(53)\end{tabular} &
  \begin{tabular}[c]{@{}c@{}}0.198\end{tabular} &
  \begin{tabular}[c]{@{}c@{}}0.296\end{tabular} &
  \begin{tabular}[c]{@{}c@{}}0.410\end{tabular} &
  \begin{tabular}[c]{@{}c@{}}0.537\end{tabular} &
  \begin{tabular}[c]{@{}c@{}}0.588\end{tabular} \\
  \hline

  \begin{tabular}[c]{@{}c@{}}Lablight(7)\end{tabular} &
  \begin{tabular}[c]{@{}c@{}}Lablight(53)\end{tabular} &
  \begin{tabular}[c]{@{}c@{}}0.276\end{tabular} &
  \begin{tabular}[c]{@{}c@{}}0.301\end{tabular} &
  \begin{tabular}[c]{@{}c@{}}0.380\end{tabular} &
  \begin{tabular}[c]{@{}c@{}}0.437\end{tabular} &
  \begin{tabular}[c]{@{}c@{}}0.488\end{tabular} \\

  \hline

  \multicolumn{7}{l}{Sunny(53)
indicates taking the samples from a sunny environment where
the distance is 53 inches.}

\end{tabular}%
}
\end{table}

\subsection{When MMW radar is dynamic}

\subsubsection{Impact of Ambience}
 We consider recognizing objects in three different environments while keeping the distance fixed. We consider taking training and testing samples from the same environment at a fixed distance. The findings are represented in Figure~\ref{fig:same_distance_cross_environment_42_84}(a) and Figure~\ref{fig:same_distance_cross_environment_42_84}(b). For 42 inches and 84 inches of the MMW radar, the recognition performance of the FCL and CNN is above 92\% when the radar is static or dynamic.
\subsubsection*{Key Takeaways}
\begin{itemize}
    \item A lablight environment (indoor) provides better recognition performance than an outdoor environment (lab environment with light and without light) when we use a FCL for a distance of 42 inches. When we apply CNN, the F1-scores are close in all environments. It is expected, as CNN is more powerful than FCL.
    
    \item When using FCL for a distance of 84 inches, the F1-score is unsatisfactory. However, the F1-score is better in sunny environments compared to other conditions. By using CNN for a distance of 84 inches, complex features can be learned and the F1-score improves from 0.754 to 0.873 in a lablight environment, as shown in Figure~\ref{fig:same_distance_cross_environment_42_84}(b). This improvement applies to other cases because radar struggles to capture sufficient information in dynamic states.
\end{itemize}

\subsubsection{Impact of Cross-Ambience}
Here, we test how data from various settings might be used to draw conclusions. This study uses the same dataset and distance for both training and testing as the previous one, but differs in that the surroundings are significantly changed. Take into account two 42-inch data sets, one from a sunny setting and the other from a lablight setting. Data that has been collected in a sunny environment is used to train the model, which is then put to the test under artificial lighting. Figure~\ref{fig:same_distance_cross_environment_42_84}(a) and Figure~\ref{fig:same_distance_cross_environment_42_84}(b) are showing the experimental findings. 
\subsubsection*{Key Takeaways}
\begin{itemize} 
\item Significant performance losses occur when the environments of the training and testing datasets are different. Figure~\ref{fig:same_distance_cross_environment_42_84}(a) and Figure~\ref{fig:same_distance_cross_environment_42_84}(b) show the performance of the deep learning models. It's to be anticipated since the distribution of data varies depending on the environment, although the distance is the same. We have used UDA and SSDA methods with 10\% and 20\% labeled data from the target domain to solve this problem. Figure~\ref{fig:same_distance_cross_environment_42_84}(a) and Figure~\ref{fig:same_distance_cross_environment_42_84}(b) illustrate that SSDA leads to a substantial improvement in performance, whereas UDA results in a comparatively smaller improvement. The F1-score rises from 0.691 to 0.898 (shown in Figure~\ref{fig:same_distance_cross_environment_42_84}(a)) after utilizing SSDA with training samples from a night environment and testing samples from a lablight environment. It is applicable to other cross-ambience situations as well.
\end{itemize}

\subsubsection{Impact of Cross-Distance}
We have observed a significant decrease in performance due to the existence of data distribution heterogeneity caused by different environments. Here, we investigate what happens to the model's F1-score when training and testing data samples have the same ambient conditions but are collected from different distances. Table~\ref{tab:same_environment_cross_height_84_42_rDNN} represents the experimental findings. 
\subsubsection*{Key Takeaways}
\begin{itemize} 
\item Similarly to cross-ambience, cross-distance leads to suboptimal outcomes. We see a similar tendency for cross-distance and cross-ambience training and testing conditions. When we apply UDA and SSDA, the performance ameliorates. The F1-score rises from 0.521 to 0.768 (shown in Table~\ref{tab:same_environment_cross_height_84_42_rDNN}) for the night environment. Here, training samples are taken from 84 inches, but testing samples are taken from 42 inches. Now, in the following section, we describe how far our future research plans have progressed.
\end{itemize}

\begin{table}[htbp]
\centering
\caption{Performance of different deep learning models
under cross distances for the same ambience settings.}
\label{tab:same_environment_cross_height_84_42_rDNN}
\resizebox{.95\linewidth}{!}{%
\begin{tabular}{@{}|c|c|c|c|c|c|c|@{}}
\hline
    
    \begin{tabular}[c]{@{}c@{}}Training\\ environment \\ 
    \end{tabular} &
    \begin{tabular}[c]{@{}c@{}}Testing\\ environment \\ 
    \end{tabular} &
    \begin{tabular}{c}F1-score (FCL) \end{tabular} &
    \begin{tabular}{c}F1-score (CNN)\end{tabular} &
    \begin{tabular}{c}F1-score \\(UDA)\end{tabular} &
    \begin{tabular}{c}F1-score \\(SSDA 10\%)\end{tabular} &
    \begin{tabular}{c}F1-score \\(SSDA 20\%)\end{tabular} \\ 
    \hline

  \begin{tabular}[c]{@{}c@{}}Sunny(84)\end{tabular} &
  \begin{tabular}[c]{@{}c@{}}Sunny(42)\end{tabular} &
  \begin{tabular}[c]{@{}c@{}}0.492\end{tabular} &
  \begin{tabular}[c]{@{}c@{}}0.514\end{tabular} &
  \begin{tabular}[c]{@{}c@{}}0.494\end{tabular} &
  \begin{tabular}[c]{@{}c@{}}0.542\end{tabular} &
  \begin{tabular}[c]{@{}c@{}}0.670\end{tabular} \\
  \hline

  \begin{tabular}[c]{@{}c@{}}Night(84)\end{tabular} &
  \begin{tabular}[c]{@{}c@{}}Night(42)\end{tabular} &
  \begin{tabular}[c]{@{}c@{}}0.571\end{tabular} &
  \begin{tabular}[c]{@{}c@{}}0.521\end{tabular} &
  \begin{tabular}[c]{@{}c@{}}0.583\end{tabular} &
  \begin{tabular}[c]{@{}c@{}}0.702\end{tabular} &
  \begin{tabular}[c]{@{}c@{}}0.768\end{tabular} \\
  \hline

  \begin{tabular}[c]{@{}c@{}}Lablight(84)\end{tabular} &
  \begin{tabular}[c]{@{}c@{}}Lablight(42)\end{tabular} &
  \begin{tabular}[c]{@{}c@{}}0.460\end{tabular} &
  \begin{tabular}[c]{@{}c@{}}0.491\end{tabular} &
  \begin{tabular}[c]{@{}c@{}}0.512\end{tabular} &
  \begin{tabular}[c]{@{}c@{}}0.603\end{tabular} &
  \begin{tabular}[c]{@{}c@{}}0.644\end{tabular} \\

 \hline

   \begin{tabular}[c]{@{}c@{}}Sunny(42)\end{tabular} &
  \begin{tabular}[c]{@{}c@{}}Sunny(84)\end{tabular} &
  \begin{tabular}[c]{@{}c@{}}0.405\end{tabular} &
  \begin{tabular}[c]{@{}c@{}}0.457\end{tabular} &
  \begin{tabular}[c]{@{}c@{}}0.461\end{tabular} &
  \begin{tabular}[c]{@{}c@{}}0.500\end{tabular} &
  \begin{tabular}[c]{@{}c@{}}0.570\end{tabular} \\
  \hline
\begin{tabular}[c]{@{}c@{}}Night(42)\end{tabular} &
  \begin{tabular}[c]{@{}c@{}}Night(84)\end{tabular} &
  \begin{tabular}[c]{@{}c@{}}0.466\end{tabular} &
  \begin{tabular}[c]{@{}c@{}}0.526\end{tabular} &
  \begin{tabular}[c]{@{}c@{}}0.543\end{tabular} &
  \begin{tabular}[c]{@{}c@{}}0.563\end{tabular} &
  \begin{tabular}[c]{@{}c@{}}0.631\end{tabular} \\
  \hline
 \begin{tabular}[c]{@{}c@{}}Lablight(42)\end{tabular} &
  \begin{tabular}[c]{@{}c@{}}Lablight(84)\end{tabular} &
  \begin{tabular}[c]{@{}c@{}}0.411\end{tabular} &
  \begin{tabular}[c]{@{}c@{}}0.422\end{tabular}&
  \begin{tabular}[c]{@{}c@{}}0.445\end{tabular} &
  \begin{tabular}[c]{@{}c@{}}0.518\end{tabular} &
  \begin{tabular}[c]{@{}c@{}}0.532\end{tabular} \\
\hline

\multicolumn{7}{l}{Sunny(42)
indicates taking the samples from a sunny environment where
the distance is 42 inches.}

\end{tabular}%
}
\end{table}

\section{Discussion}
We address the issues of developing and implementing a deep learning model that utilizes MMW radar. These include potential noise in the dataset, challenges in real-time implementation with point cloud radar data, and the system's potential impact on daily life.

\begin{itemize}
    \item During the data collection process, surrounding objects can potentially introduce noise into the dataset. Although such objects may be considered noise, we believe that our model is robust enough to recognize actual objects in the presence of noise. However, in the future, we may be able to improve the performance of our model by reducing the noise introduced by these surrounding objects.
    
    \item Integrating the object recognition pipeline into a real-time system can be straightforward, but there are various implementation challenges to consider, such as power consumption and noise detection from surrounding objects. These challenges present concrete obstacles to achieving real-time implementation.
    
    \item Our deep learning methodology has exclusively utilized the processed sparse point cloud data generated from the MMW radar. If we incorporate divergent types of MMW radar data, it could lead to processing and computational difficulties, potentially requiring significant alterations to the hyper-parameters of our model.
    
    \item The potential use of MMW radar technology to recognize both familiar and unfamiliar objects, activities, and health issues concurrently may have a profound influence on daily life. Given the limitations on camera placement due to privacy concerns, radar may serve as an excellent substitute for identifying criminal or other daily activities.        
     
\end{itemize}

We are nearing the conclusion of our investigations, where we summarize our current constraints and future research scopes.
\vspace{-1ex}
\section{Conclusion}

\label{sec:conclusion}


Our study has shown promising results for static object recognition using MMW radar. We have evaluated the performance of our proposed model under various conditions and found that it performs well on data with a homogenous ambiance but poorly on data with a heterogeneous ambiance. To address this issue, we have introduced domain adaptation techniques to improve accuracy. However, our current model only focuses on recognizing static objects, and to address real-world applications, we need to develop more advanced techniques for recognizing known or unknown objects, activities, and health issues simultaneously. This technology has diverse practical applications, especially in areas where privacy issues constrain the deployment of cameras, such as security, surveillance, and monitoring.
\vspace{-1ex}
\section*{ACKNOWLEDGEMENT}
This research is supported by the U.S. Army Grant \#W911NF2120076, and NSF Research Experience for Undergraduates (REU) grant \#CNS-2050999. We want to thank Dr. Kelly Sherbondy for his collaboration on this project.

\bibliographystyle{unsrt}
\bibliography{main}
\end{document}